\RequirePackage{fix-cm}

\documentclass[twocolumn]{svjour3}           

\smartqed  

\usepackage[hyphens]{url}
\usepackage{graphicx}
\usepackage{amsmath,amssymb,amsfonts}
\usepackage{algorithmic}
\usepackage{graphicx}
\usepackage{textcomp}
\usepackage[svgnames,table]{xcolor}
\usepackage{float} 
\usepackage{lipsum}
\usepackage{amsmath}
\usepackage{lipsum}
\usepackage{multirow}
\usepackage[utf8]{inputenc}
\usepackage[english]{babel}
\usepackage{booktabs}
\usepackage{caption}
\usepackage{pgfplots}
\usepackage{subcaption}
\usepackage{makecell}
\usepackage[misc]{ifsym}
\usepackage{parskip}

\usetikzlibrary{shadows}
\usetikzlibrary{decorations.pathreplacing}

\def\BibTeX{{\rm B\kern-.05em{\sc i\kern-.025em b}\kern-.08em
    T\kern-.1667em\lower.7ex\hbox{E}\kern-.125emX}}
    
\usepackage{import}
\usetikzlibrary{quotes,arrows.meta}
\usetikzlibrary{positioning}

\def\edgecolor{rgb:blue,4;red,1;green,4;black,3}
\newcommand{\midarrow}{\tikz \draw[-Stealth,line width =0.8mm,draw=\edgecolor] (-0.3,0) -- ++(0.3,0);}

\usepackage{Ball}
\usepackage{Box}
\usepackage{RightBandedBox}

\let\bf\bfseries

\newcommand{\cb}{\cellcolor{black!20}\bfseries}

\usetikzlibrary{positioning}

\def\ConvColor{rgb:yellow,5;red,2.5;white,5}
\def\ConvReluColor{rgb:yellow,5;red,5;white,5}
\def\PoolColor{rgb:red,1;black,.3}
\def\FcColor{rgb:blue,5;red,2.5;white,5}
\def\FcReluColor{rgb:blue,5;red,5;white,4}
\def\SoftmaxColor{rgb:magenta,5;black,7}

\usepgfplotslibrary{colorbrewer}
\pgfplotsset{compat=1.6}

\usepackage[compact]{titlesec}
\usepackage{hyperref}
\hypersetup{colorlinks=true,breaklinks=true}

\begin{document}

\title{Incremental Class Learning using Variational Autoencoders with Similarity Learning}
\titlerunning{Incremental Similarity Learning using VAEs}

\author{Jiahao Huo\textsuperscript{1}(\Letter) \and
        Terence L. van Zyl\textsuperscript{2}
}

\authorrunning{Jiahao Huo \and Terence L. van Zyl}

\institute{Jiahao Huo\textsuperscript{1}(\Letter) \at
           \textit{Institute for Intelligent Systems} \\
           \textit{University of Johannesburg}, South Africa \\
           \email{216045414@student.uj.ac.za}           
           \and
           Terence L. van Zyl\textsuperscript{2} \at
           \textit{Institute for Intelligent Systems} \\
           \textit{University of Johannesburg}, South Africa \\
           \email{tvanzyl@uj.ac.za}
}

\date{Received: date / Accepted: date}

\maketitle

\begin{abstract}
Catastrophic forgetting in neural networks during incremental learning remains a challenging problem. Previous research investigated catastrophic forgetting in fully connected networks, with some earlier work exploring activation functions and learning algorithms. Applications of neural networks have been extended to include similarity learning. Understanding how similarity learning loss functions would be affected by catastrophic forgetting is of significant interest. Our research investigates catastrophic forgetting for four well-known similarity-based loss functions during incremental class learning. The loss functions are Angular, Contrastive, Center, and Triplet loss. Our results show that the catastrophic forgetting rate differs across loss functions on multiple datasets. The Angular loss was least affected, followed by Contrastive, Triplet loss, and Center loss with good mining techniques. We implemented three existing incremental learning techniques, iCaRL, EWC, and EBLL. We further proposed a novel technique using Variational Autoencoders (VAEs) to generate representation as exemplars passed through the network's intermediate layers. Our method outperformed three existing state-of-the-art techniques. We show that one does not require stored images (exemplars) for incremental learning with similarity learning. The generated representations from VAEs help preserve regions of the embedding space used by prior knowledge so that new knowledge does not ``overwrite'' it.
\keywords{Catastrophic forgetting \and incremental learning \and similarity learning \and convolutional neural network (CNN)}
\end{abstract}

\section{Introduction}
\label{intro}

Catastrophic forgetting in deep neural networks remains an open challenge~\cite{goodfellow2013empirical,fernando2017pathnet,robins1995catastrophic,draelos2017neurogenesis}. Some machine learning algorithms cannot retain prior knowledge while incrementally learning and suffer from catastrophic forgetting. Incremental learning refers to updating a model as new data becomes available or extending the model to incorporate additional tasks. This catastrophic forgetting occurs during training in which new data or tasks are presented to the model with few or no examples drawn from prior learned distributions~\cite{mccloskey1989catastrophic,ratcliff1990connectionist}. Deep learning is widely used in the field of computer vision and is shown to have many real-world applications~\cite{wang2016hierarchical,wang2020sa,wang2021unsupervised,wang2021learning}. Historically, the focus has been on incremental supervised classification. Similarity learning has emerged as a significant area of machine learning research with many real-world applications~\cite{dlamini2019author,van2020unique,manack2020deep,burns2020automated}. Most current research focuses on preserving exemplars of previous classes to remind the network of past knowledge by finding exemplars that are useful for knowledge retention~\cite{rebuffi2017icarl,buzzega2020dark,shim2021online}. Recent advances include techniques such as using supervised contrastive learning with nearest-class-mean to maximize the use of exemplars~\cite{mai2021supervised}. In addition, some recent studies have used generative models and knowledge distillation techniques to generate synthetic data and update the memory bank periodically during incremental learning~\cite{binici2022preventing}. However, there is a lack of empirical evidence regarding the extent to which similarity learning is affected by catastrophic forgetting. Further, few techniques exist to solve catastrophically forgetting in similarity learning~\cite{9311580}. As such, there is a need to investigate the viability of existing solutions aimed at reducing catastrophic forgetting during incremental similarity learning, which is fundamentally different from incremental learning for classification.

Given the context of catastrophic forgetting during incremental similarity learning, we introduced a novel method using simple Variational Autoencoders (VAEs). The method generates representations to supplement previously seen data by extracting them after the convolutional layers. We further implement three state-of-the-art solutions to reduce catastrophic forgetting during incremental learning. The prior solutions consist of Elastic Weight Consolidation (EWC) \cite{kirkpatrick2017overcoming}, Encoder-Based Lifelong Learning (EBLL) \cite{rannen2017encoder}, and incremental Classifier and Representation Learning (iCaRL) \cite{rebuffi2017icarl}. This paper compares the impact of catastrophic forgetting in incremental similarity learning for our method contrasted against the existing methods.

In summary, our contributions are:
\begin{itemize}
    \item We show that our method outperformed the three existing solutions for mitigating catastrophic forgetting during incremental similarity learning of new classes.
    \item We demonstrate that the generated representations from VAEs work as well as images for exemplars.
    \item We highlight that with good mining techniques, Angular loss is least affected by catastrophic forgetting.
    \item We reinforce that more challenging tasks and an increased number of initial classes result in increased catastrophic forgetting.
\end{itemize}

\section{Background} 

\subsection{Catastrophic Forgetting in Deep Neural Networks}

Previously, Goodfellow \emph{\textit{et al.}} \cite{goodfellow2013empirical} investigated catastrophic forgetting in gradient-based neural networks used for classification. Catastrophic forgetting causes a convolutional neural network classifier to lose representations of old data resulting in a loss of accuracy on old data \cite{kemker2018measuring}. These results showed that various combinations of activation function and learning were affected differently by catastrophic forgetting. 

Later, work by Thanh-Tung \emph{\textit{et al.}} \cite{thanh2018catastrophic} conducted an in-depth analysis of catastrophic forgetting in Generative Adversarial Networks (GAN). The study focused on the causes and effects of catastrophic forgetting and how they relate to mode collapse and non-convergence of the GANs. The outcome highlighted that GANs suffer from catastrophic forgetting even when trained on datasets generated from a mixture of eight Gaussian models. Further analysis of incremental learning of real-world image datasets such as CIFAR-10 and CelebA exhibited the same limitations. The challenges arise from the previous information learned not being used for the new task. Second, the current task is too different from previous tasks; therefore, knowledge is not reused and overwritten.

Existing work by Seff \emph{\textit{et al.}} \cite{seff2017continual} demonstrated the use of EWC \cite{kirkpatrick2017overcoming} to overcome catastrophic forgetting in GANs during sequential training on a set of distributions. The study by Choi \emph{\textit{et al.}} \cite{choi2019autoencoder} proposed the use of an autoencoder-based incremental class learning method for classification. Catastrophic forgetting in an autoencoder is when the network losses its ability to reconstruct old data when learning new data. The method functioned without a softmax classification layer, as typical in conventional classification models. The method is premised on a metric-based classification method, nearest-class-mean (NCM), by Mensink \emph{\textit{et al.}} \cite{mensink2013distance}. The main idea is to use a pre-trained fixed network as a feature extractor for the autoencoder. The autoencoder is then trained on these feature embeddings. The encoded representations from the resulting encoder are used for cosine similarity-based classification. The problem of catastrophic forgetting occurs when the autoencoder is fine-tuned for feature embedding for new incrementally learned classes. To overcome this difficulty, the authors use regularization techniques: Synaptic Intelligence (SI) \cite{zenke2017continual} and Memory Aware Synapses (MAS) \cite{aljundi2018memory}. The techniques add a term to the existing loss function during the incremental class learning phase. The methods demonstrated good memory retention without the need to train on older data.

\subsection{Reducing Catastrophic Forgetting}
\label{previouswork}

\subsubsection{Elastic Weight Consolidation (EWC)}

Elastic Weight Consolidation is a method proposed by Kirkpatrick \emph{\textit{et al.}} \cite{kirkpatrick2017overcoming} aimed at overcoming the limitations of catastrophic forgetting in neural networks during incremental learning. EWC selectively slows down the learning of weights important to previously learned tasks. The constraint to slow down weight updates is a Gaussian distribution modelled using the network weights as the mean and the diagonal of the Fisher information matrix from the previous task. The update constraint is shown as follows:
\begin{equation}
    L(\theta) = L_{t}(\theta) + \sum_{i}\frac{\lambda}{2}F_{i}(\theta_{i} - \theta_{t-1,i}^{*})^2 , 
\end{equation}
where $L(\theta)$ is the combined loss. $\theta$ are the network parameters. $L_{t}(\theta)$ is loss of the current training session at time $t$. $\lambda$ is a hyper-parameter that indicates the importance of the old tasks compared to the new tasks. $i$ represents each parameter of the network. $F$ is the Fisher Information Matrix used to constrain the important weights for previously learned tasks to the original values. $\theta_{t-1}^{*}$ are the trainable parameters from the previously trained model of a training session, $t-1$. They compute the Fisher Information Matrix using the gradient of the cross entropy loss from the network's output. 

\subsubsection{Incremental Classifier and Representation Learning (iCaRL)}

Incremental Classifier and Representation Learning is a method proposed by Rebuffi \emph{\textit{et al.}} \cite{rebuffi2017icarl} for reducing catastrophic forgetting.  iCaRL learns a strong classifier and data representations simultaneously. As a result, it is compatible with deep neural networks. iCaRL primarily relies on the storing of exemplars from previously seen classes. Each class's exemplar set is constructed by storing $k$ images ranked by the closeness of their representation to the class's mean representation. This selection of the $k$ closest images is known as the herding algorithm. The stored exemplars supplement the incremental learning phase of new classes using knowledge distillation. Classification is performed using the stored exemplars following the nearest-mean-of-exemplars rule: \textit{A new image is classified to the class of the exemplar closest to it}. iCaRL is shown to learn classes incrementally over a longer period than other methods, which failed more quickly. 

\subsubsection{Encoder-Based Lifelong Learning (EBLL)}

Encoder-Based Lifelong Learning is proposed by Rannen \emph{\textit{et al.}} \cite{rannen2017encoder} for incremental learning in classification tasks. The authors demonstrated the problem of catastrophic forgetting in deep convolutional neural networks (DCNN) AlexNet. The study highlighted a classification performance drop on a previously learned task when a DCNN is fined-tuned for newer classification tasks. The authors proposed using lightweight autoencoders to preserve the feature representations learned by the base network (AlexNet) that were optimal for the task before learning the next task. An autoencoder is trained after the network learns each new task, increasing storage requirements for each autoencoder. The methods significantly reduced the catastrophic forgetting that occurred when incrementally learning new tasks for classification.

\section{Methodology}

\subsection{Proposed Approach}

We propose a novel approach that bridges the ideas of Rannen \emph{\textit{et al.}} \cite{rannen2017encoder} and Rebuffi \emph{\textit{et al.}} \cite{rebuffi2017icarl}. Rannen \emph{\textit{et al.}} \cite{rannen2017encoder} constrain the weights of the feature extraction layers (convolutions) that were optimal for previous tasks with an autoencoder. The solution effectively reuses the feature extraction layers (convolutions) on new tasks. Each task is tested independently from the others with its classification layer. The approach yields promising results by finding a middle ground across tasks.

The method iCaRL by Rebuffi \emph{\textit{et al.}} \cite{rebuffi2017icarl} largely depends on the storage and usage of exemplars. As reported, the performance of iCaRL decreases with time as the number of exemplars per class is reduced to accommodate new classes. Eventually, the stored exemplars may not be sufficient to represent all classes.

In our approach, we train a new variational autoencoder (VAE) for each class. The VAEs learn representations from the end of the convolutional layers. VAEs allow us to sample representations of previously seen classes as an output from the convolutional layers as an alternative to reconstructing the entire image. A complete CNN approach would be more computationally expensive and require significantly more complex VAEs, but it is not infeasible. 

Our method requires that the convolutional layers be frozen after their initial training. Alternatively, pre-trained frozen convolutional layers from a base/foundational model could be used, as is common in transfer learning~\cite{liu2021transtailor}. The convolutional layers are frozen since the reconstructions from the VAEs will not match if the weights in the convolutional layers change. The VAEs generate samples from previously seen classes combined with the new classes during incremental class training to perform incremental similarity learning. The autoencoder's reconstruction loss function varies depending on the network's last convolutional layer's activation function. For example, in our case, the last convolutional layers use sigmoid activation. Therefore we used the Binary Cross-Entropy objective function to calculate the reconstruction errors summed with the Kullback-Leibler divergence. The loss function to update the VAEs is given as follows:
\begin{equation}
\begin{split}
    L_{VAE} = -\frac{1}{N}\sum_{i=1}^{N}y_{i}\cdot \log(p(y_i)) + (1-y_{i})\cdot \log(1-p(y_{i})) \\ + \frac{1}{2}(\exp(\sigma^2)+\mu^2 -1 - \sigma^2),
\end{split}
\end{equation}
where $\sigma^2$ is the variance of the full dataset  and $\mu$ is the mean. The first term is the Binary Cross-Entropy reconstruction loss, and the second is the Kullback–Leibler divergence. $N$ is the length of the representation vector.  $y_{i}$ is the i-th value in the representation vector. $p(y_{i})$ is the probability of the value $y_{i}$.

We use the angle-wise distillation loss on the generated examples from the VAEs~\cite{park2019relational}. This is used when updating the network during incremental learning, similar to the approach by iCaRL~\cite{rebuffi2017icarl}. The angle-wise distillation loss is defined as follows:
\begin{equation}\label{eqn:distil}
L_{A} =\sum_{x_{i},x_{j},x_{k}\in X^{3}} \ell_{h}(V_{A}(t_{i},t_{j},t_{k}),V_{A}(s_{i},s_{j},s_{k})), 
\end{equation}
where $V_{A}$ is the angle-wise potential between the triplet images, $x_{i},x_{j},x_{k}$ and $\ell_{h}$ refers to the Huber loss. $t_{i},t_{j},t_{k}$ is the output of the teacher network (model that is trained and frozen) for the triplet images. $s_{i},s_{j},s_{k}$ is the output of the student network (model that is being updated). The loss penalizes the Angular differences between the two networks.

Our incremental learning setup makes use of the following loss function during the incremental step:
\begin{equation}
    L_{\text{incremental}} = l_\text{similarity learning} + \lambda_{\text{distil}}*L_{A},
\end{equation}
where $l_\text{similarity learning}$ are the Contrastive, Angular and Triplet loss functions shown in Section \ref{section:losses}. $L_{A}$ is the angle-wise distillation for similarity learning. $\lambda_{\text{distil}}$ is the importance placed on the angle-wise distillation loss. The student output is the output at each new step from the last layer of the network being incrementally trained. The teacher output is the output from the last layer of the frozen network before performing each new incremental train step. 

We were required to make use of an alternative to the original distillation loss\~cite{hinton2015distilling} when using Center loss, defined as:
\begin{equation}
    L_{D}(t_{o},s_{o}) = -\sum_{i=1}^{l}t_{o}^{i}log(s_{o}^{i}),
\end{equation}
where $l$ is the number of labels, $t_{o}^{i}$ and $s_{o}^{}i$ are the modified versions of the teacher model outputs and the current student model outputs. As a result, the modified loss function during the incremental step for Center loss is defined as:
\begin{equation}
    L_{\text{incremental}} = l_\text{similarity learning} + \lambda_{\text{distil}}*L_{D},
\end{equation}
where $L_{D}$ is defined above, and the rest remains the same.

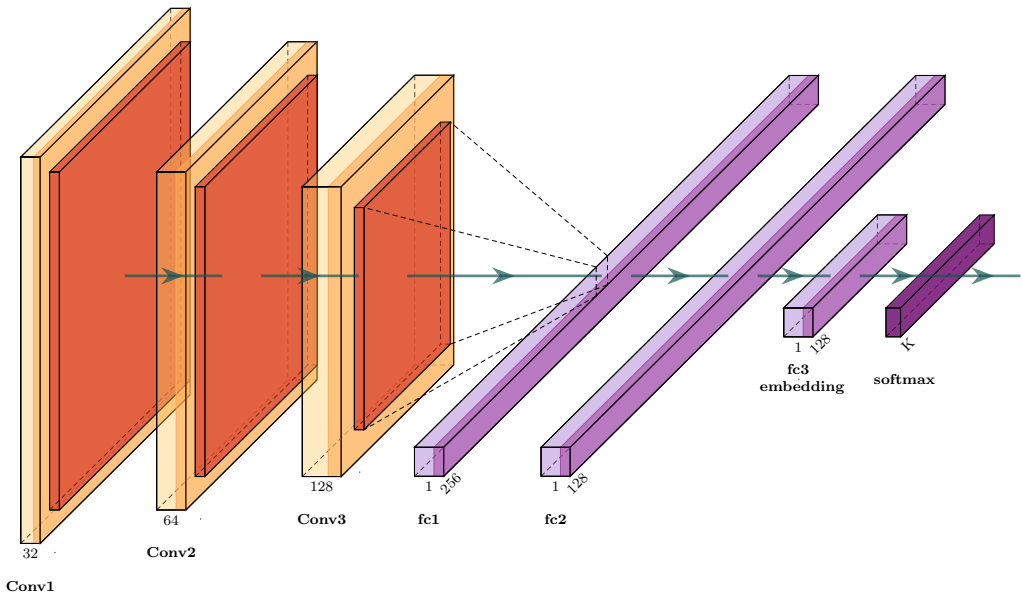
\begin{figure*}[ht!]
\centering
\resizebox{0.8\textwidth}{!}{%
    \begin{tikzpicture}
    \tikzstyle{connection}=[ultra thick,every node/.style={sloped,allow upside down},draw=\edgecolor,opacity=.7]
    \pic[shift={(0,0,0)}] at (0,0,0) {RightBandedBox={name=cr1,caption=Conv1,%
            xlabel={{"32",,}},fill=\ConvColor,bandfill=\ConvReluColor,%
            height=40,width={2},depth=40}};
    \pic[shift={(0,0,0)}] at (cr1-east) {Box={name=p1,caption={},%
            fill=\PoolColor,opacity=.5,height=35,width={1},depth=35}};
    \pic[shift={(2,0,0)}] at (p1-east) {RightBandedBox={name=cr2,caption=Conv2,%
            xlabel={{"64",,}},fill=\ConvColor,bandfill=\ConvReluColor,%
            height=35,width={3},depth=35}};
    \pic[shift={(0,0,0)}] at (cr2-east) {Box={name=p2,caption={},%
            fill=\PoolColor,opacity=.5,height=30,width={1},depth=30}};
    \pic[shift={(2,0,0)}] at (p2-east) {RightBandedBox={name=cr3,caption=Conv3,%
            xlabel={{"128",,}},fill=\ConvColor,bandfill=\ConvReluColor,%
            height=30,width={4},depth=30}};
    \pic[shift={(0,0,0)}] at (cr3-east) {Box={name=p3,caption={},%
            fill=\PoolColor,opacity=.5,height=23,width={1},depth=23}};
    \pic[shift={(4,0,0)}] at (p3-east) {RightBandedBox={name=fc6,caption=fc1,%
            xlabel={{"1",""}},zlabel=256,fill=\FcColor,bandfill=\FcReluColor,%
            height=3,width={3},depth=100}};
    \pic[shift={(2,0,0)}] at (fc6-east) {RightBandedBox={name=fc7,caption=fc2,%
            xlabel={{"1","dummy"}},zlabel=128,fill=\FcColor,bandfill=\FcReluColor,%
            height=3,width={3},depth=100}};
    \pic[shift={(1.5,0,0)}] at (fc7-east) {RightBandedBox={name=fc8,caption=fc3\\embedding,%
            xlabel={{"1","dummy"}},zlabel=128,fill=\FcColor,bandfill=\FcReluColor,%
            height=3,width={3},depth=25}};
    
    \pic[shift={(1.5,0,0)}] at (fc8-east) {Box={name=softmax,caption= softmax,%
            xlabel={{"","dummy"}},zlabel=K,opacity=.8,fill=\SoftmaxColor,%
            height=3,width={1.5},depth=25}};
        
    \draw [connection]  (p1-east)        -- node {\midarrow} (cr2-west);
    \draw [connection]  (p2-east)        -- node {\midarrow} (cr3-west);
    \draw [connection]  (p3-east)        -- node {\midarrow} (fc6-west);
    \draw [connection]  (fc6-east)       -- node {\midarrow} (fc7-west);
    \draw [connection]  (fc7-east)       -- node {\midarrow} (fc8-west);
    \draw [connection]  (fc8-east)   -- node {\midarrow} (softmax-east);
    \draw [connection]  (softmax-east)   -- node {\midarrow} ++(1.5,0,0);

    \draw[densely dashed]
        (fc6-west)++(0, 1.5*.2, 1.5*.2) coordinate(a) -- (p3-nearnortheast)
        (fc6-west)++(0,-1.5*.2, 1.5*.2) coordinate(b) -- (p3-nearsoutheast)
        (fc6-west)++(0,-1.5*.2,-1.5*.2) coordinate(c) -- (p3-farsoutheast)
        (fc6-west)++(0, 1.5*.2,-1.5*.2) coordinate(d) -- (p3-farnortheast)
        (a)--(b)--(c)--(d);
    \end{tikzpicture}
}
\caption{Architecture of our convolution neural network. The yellow layers represent the convolutions, the orange represents the pooling layers, and the purple is fully connected layers. The K refers to the number of classes in the data. Note that the softmax fully connected layer is only used for Center loss. For the rest of the loss functions, we only use the network up to fc3.}
\label{fig:Architecture}
\end{figure*}

For our method, we experiment with a ResNet9, VGG11 and a simple CNN backbone. The ResNet9 and VGG11 evaluate performance on commonly known deep neural network architectures. The simple CNN backbone, shown in Figure \ref{fig:Architecture}, is a neural network with three 2D convolutional layers interleaved with a max-pooling layer. The final max-pooling layer is followed by a Flatten layer and two ReLU-activated fully connected layers. The output layer is a linearly activated, fully connected layer with an output size of $128$. 

\begin{figure*}[ht!]
  \centering
  \includegraphics[width=.8\textwidth]{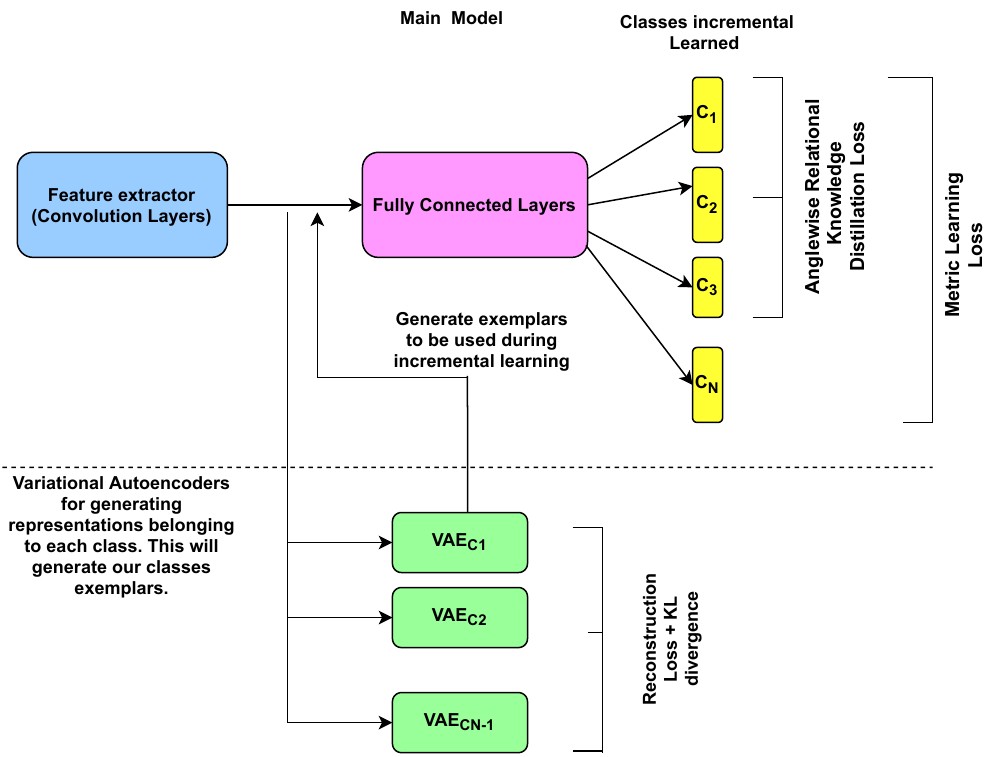}
  \caption{A diagram demonstrating how our approach works. Feature extractor represents the convolutional layers of the network. After initially training on the base-training set, the feature extractor is frozen. The VAEs can then generate consistent representations for each class passed through.}
  \label{fig:ourapproach}
\end{figure*}

We ensured that all networks were trained to perform reasonably well on all datasets so that we could observe the effects of catastrophic forgetting. We change the activation function layer of the last 2D convolutional layer from ReLU to sigmoid to use binary cross-entropy as our reconstruction loss for our VAEs. Our VAE architecture consisted of 512 input layers (the same size as out after the convolution layers), followed by 256, 128, and 128 for the bottleneck. Then these layers are reconstructed symmetrically in reverse.  Figure \ref{fig:ourapproach} illustrates our approach.

\subsection{Loss Functions}
\label{section:losses}

The Previous research mentioned in Section \ref{previouswork} has yet to be investigated in the similarity learning context to the best of our knowledge. Similarity learning is an area of supervised learning that explores the similarity between objects using a distance metric~\cite{chechik2010large}. In this paper, we consider four prominent loss functions used in similarity learning: 

\subsubsection{Triplet Loss} 

Triplet loss by Wang \emph{\textit{et al.}} \cite{wang2014learning,schroff2015facenet} has been shown to learn good feature representations for determining image and video similarity~\cite{huo2020unique}. The triplet comprises an anchor ground truth image, a positive and negative image. The positive image belongs to the same identity as the ground truth, and a negative image is selected from an identity that differs from the anchor. The loss is given as follows:
\begin{equation}
    \mathcal{L} = \max(d(a,p) - d(a,n) + \text{margin},0),
\end{equation}
where $d$ represents euclidean distance, $a$ is the anchor ground truth image, $p$ is the positive image, $n$ is the negative image. The margin represents a radius around the anchor and determines the degree to which the negative image is pushed away. The function optimizes the distance between the anchor-positive and anchor-negative simultaneously by bringing the positive pair closer and pushing the negative pairs apart.

\subsubsection{Contrastive Loss} 

Contrastive loss finds optimal features by using positive and negative non-matching pairs of images. The function is given as follows:
\begin{equation}
\begin{aligned}
\mathcal{L} = & \frac{1}{N_{P}}\sum_{(a,b) \in P} \left( \frac{1}{2}(1-Y_{(a,b)})(d(\tilde{a}, \tilde{b}))^2 \right.  \\  
 & \left. +\frac{1}{2}(Y_{(a,b)})\{\max(0,\text{margin}-d(\tilde{a}, \tilde{b}))\}^2\right),
\end{aligned}
\end{equation}
where $Y_{(a,b)}$ represents the label 0 or 1 and is 0 if the input image pair $(a,b)$ are from the same class and 1 otherwise. P is the set of all image pairs. $N_{P}$ is the number of image pairs in the P set. $d(\tilde{a}, \tilde{b})$ represents the Euclidean distance between the output feature representations of the network for the pair of images $(a,b)$. The loss function differs from Triplet loss in that it tries to minimize the distance between positive pairs and maximize negative pairs in separate steps.

\subsubsection{Angular Loss} 

Wang \emph{\textit{et al.}} \cite{wang2017deep} aimed to overcome some of the challenges with Triplet loss. These problems include the infeasibility of exhaustive sampling and using a single global margin $m$. Angular loss addresses these by constraining the angles of the triangle between the triplets. The resulting rotation and scale invariance make the loss function more robust to significant feature variations. The Angular loss of a batch $\mathcal{B}$ of size $N$ is defined as:
\begin{equation}
\begin{aligned}
    l_{ang}(\mathcal{B}) &= \frac{1}{N}\sum_{x_{a}\in \mathcal{B}}\{\log[ 1 +\sum_{\substack{x_{n}\in \mathcal{B} \\ y_n \not= y_a,y_p}} \exp(f_{a,p,n}) ]\}   , 
\end{aligned}
\end{equation}
where $x_a$ is the anchor image. $x_n$ is a negative image (different class from anchor). Function $f$ is defined as
\begin{equation}
    f_{a,p,n} = 4\tan^{2}\alpha(x_a+x_p)^{T}x_n - 2(1+\tan^{2}\alpha)x_{a}^{T}x_p ,
\end{equation}
where $x_p$ is a positive image (same class as anchor image). $\alpha$ is the degree of the angle. $y_n$ is the class label of the negative image. $y_a$ is the class label of the anchor image. $y_p$ is the class label of the positive image.

\subsubsection{Center Loss} 

Wen \emph{\textit{et al.}} \cite{wen2016discriminative} try to enhance the discriminative power of learned features specifically for facial recognition. The loss function learns a centre for the features of each unique class. It simultaneously penalizes the distances between the features of the images and their corresponding class centres that maximize inter-class separation and intra-class compactness. Center loss cannot be used directly as a loss function and is therefore paired with softmax as defined by:
\begin{equation}
    L_{\text{s + c}} = -\sum_{i=1}^{m}\log \frac{e^{W^{T}_{y_{i}}x_i + b_{y_i}}}{\sum_{j=1}^{n} e^{W^{T}_{y_{j}}x_{i} + b_{y_{j}}}} + \frac{\lambda}{2}\sum_{i=1}^{m}\lVert x_i - c_{y_i}\rVert_{2}^{2} ,
\end{equation}
where $L_{\text{s + c}}$ denotes the loss comprising of both softmax and Center loss. The left term is the formula for the softmax function. The right term is the formula for Center loss. $x_i$ denotes the features from the network. $c_{y_i}$ denotes the centre value for class $y_{i}$. $W^T_{y_{i}}$ denotes the weight of the last fully connected layer corresponding to the $y_{i}$th class, and $b_{y_{i}}$ is the $y_{i}$th class bias. $W^T_{y_{j}}$ is column of the weights corresponding to the $y_{j}$th class and $b_{y_{j}}$ is the $y_{j}$th class bias. $m$ and $n$ are the numbers of classes. $\lambda$ is a scalar representing the importance of weighting between the two losses.

\subsection{Datasets} 

To analyze the impact of catastrophic forgetting in similarity learning, all methods are subjected to incremental learning scenarios on well-known datasets. The datasets used are the MNIST, EMNIST, FashionMNIST (FMNIST), and CIFAR10. MNIST is handwritten digits~\cite{lecun2010mnist}. EMNIST is upper and lower case letters from the English alphabet~\cite{cohen2017emnist}. We regard the upper/lower cases as coming from the same class to increase the difficulty. Fashion-MNIST obtained images of clothes and shoes from Zalando's articles~\cite{xiao2017/online}. CIFAR10 is low-resolution objects~\cite{Krizhevsky09learningmultiple}. The MNIST classes are unbalanced, but the sample sizes for each class are very similar. For all other datasets, the classes are balanced. We used the Scikit-learn\cite{scikit-learn} stratified split to take $20\%$ from the training dataset to be used as validation data during training.

Preprocessing done on the MNIST and EMNIST datasets comprised normalizing the pixel values with the mean and standard deviation of [0.1307,0.3801], respectively. For FashionMNIST, we normalize the pixel values with mean and standard deviation values of  [0.2860, 0.3530], respectively. On CIFAR10, the pixels were normalized with the mean and standard deviation values of [(0.4914, 0.4822, 0.4465), (0.2023, 0.1994, 0.2010)], respectively. Additionally, padding of size four and data augmentation through horizontal flipping during training was used. The padding and augmentation are used to overcome the datasets' toughness and, in so doing, obtain sufficient performance to demonstrate the effects of catastrophic forgetting.

\subsection{Experiment Setup}

\subsubsection{Pair and Triplet Mining}

Pairs and triplets for training and validation were generated online during training. We mined triplet images with semi-hard negative online mining on positive and negative image pairs. A variety of diverse classes during training allows us to optimise distances between positive and negative classes. We performed pair margin mining to generate pairs of images for Contrastive loss. An Angular miner generated triplets at an angle greater than $\alpha$. All mining was performed using the PyTorch Metric Learning library \cite{musgrave2020pytorch} with the hyper-parameters specified in Section~\ref{section:hyperparam}.

\subsubsection{Hyper-parameters} 
\label{section:hyperparam} 

The angle, $\alpha$, for the Angular loss in our implementation was $45^\circ$, previously found as optimal for object detection\cite{wang2017deep} for MNIST and EMNIST datasets. An angle of $[35,40]$ was used for CIFAR10 and FashionMNIST, obtained by grid-search on a range between $30$ and $50$ in steps of $5$. The margin for Contrastive and Triplet loss functions were $1.0$ and $1.25$, respectively. The margins were determined through grid-search on the range $0.25$ to $2.0$ in steps of $0.25$. All grid searches were done using the validation set from FashionMNIST, MNIST, and EMNIST. Due to low accuracy on CIFAR10, a manual hyper-parameter search was performed using common values from the literature. We found that the margins for Contrastive ($0.3841$) and Triplet ($0.0961$) loss from Mugrave \emph{\textit{et al.}} \cite{musgrave2020metric} worked best. The hyper-parameters [$\lambda$, $\alpha$] for Center loss were [$1.0$, $0.5$] respectively previously shown to have good results for class separation~\cite{wang2017deep}. We used an importance value of 150 for $\lambda$ across all the experiments for EWC. We weighted the similarity learning loss and distillation loss equally by multiplying both by the $1.0$ for iCaRL and our approach. The similarity learning loss and mean square error loss of the embeddings were weighted when summed during CNN and autoencoder update in the EBLL approach with a $\lambda$ value of $1.0$.

\subsubsection{Exemplars for iCaRL}

Rebuffi \emph{\textit{et al.}} \cite{rebuffi2017icarl}, used $2000$ exemplars for CIFAR100, which results in an average ratio of 20 images per class. Therefore, in our experiments, we limited the number of exemplars for MNIST, FashionMNIST, and CIFAR10 to 200. For EMNIST letters, we limited the number of exemplars to 520 for the 26-letter classes. 

\subsubsection{Training and Testing}

We used an almost identical procedure for incremental learning to that of Kemker \emph{\textit{et al.}} \cite{kemker2018measuring}. We start with half ($5$) of the classes from each dataset: MNIST, FashionMNIST, and CIFAR10. For EMNIST, half of the classes are $13$. Subsequent training contains data from a single new unseen class. A high-level overview of the steps followed are:
\begin{enumerate}
    \item We take all the classes and split them into two sets of classes. A baseline set and the incremental set of classes.
    \item We split the baseline set into base-training, base-validation and base-test data sets.
    \item We split the incremental set into inc-training, inc-validation and inc-test data sets.
    \item We use the base-training and base-validation set to train our initial base models for incremental learning.
    \item We take one unseen class from our incremental set of classes and the previously seen class to supplement the unseen class.
    \item We retrain our base model with the inc-training data set for that unseen class.
    \item We use the base-test data set from our baseline set to record the mAP@R after each step.
    \item We repeat from step 5 until all of the incremental sets' classes are exhausted.
\end{enumerate}

Since similarity learning loss functions require at least two classes, we pair a single class from the previously learned data with the new class. We measure the mean average precision (mAP@R) on the new class after training to assess if the models are still learning. All the models were trained for a maximum of $50$ epochs, and the best models were selected using early stopping on the validation set. The Adam optimizer was employed with a learning rate of $0.001$, $\beta_1$ value of $0.9$, and $\beta_2$ value of $0.999$. For our method, we trained one variational autoencoder for each class the network has seen for each incremental training step. The same Adam optimizer was used for training. The EBLL method trained one autoencoder after each incremental class learning step using the same Adam optimizer specified above.

We randomly split the data into two sets of classes consisting of a baseline set and an incremental set. By doing this, we can get average results for different combinations of class splits of incremental learning. We repeat the experiment ten times for each incremental learning method model on each dataset (total: $800$) using randomly seeded baseline and incremental splits as previously specified. Each run consisted of the same size training and validation split, while the base-test and inc-test sets remained the same for each incremental learning method to keep results consistent. 

The models' output is a feature representation of size $128$ per image evaluated using mean Average Precision at $R$ (mAP@R). Average precision at $R$ (AP@$R$) is calculated using a query to retrieve the top $R$ related images from the database. The AP at $R$ is given by: 
\begin{equation}\label{eqn:MAP}
    \text{AP@}R = \frac{1}{\text{R}}\sum_{k=1}^{R} \text{P@}k \times \text{Rel@}k ,
\end{equation}
where $R$ is the total number of images in the database belonging to the same class as the query image. P@$k$ refers to the precision at $k$, and Rel@$k$ is a relevance score which equals 1 if the document at rank $k$ is relevant and equals 0 otherwise. mAP is the average of the AP over all possible image queries and $k$ up to $R$. $R$ is the total number of images belonging to the same class as the query.

\subsection{Hardware and Software}

We used two machines. An AMD RYZEN 3600 processor, 16 GB of RAM, and a GTX 1060 6GB GPU and an AMD RYZEN 3700X processor, 16 GB of RAM and an RTX 2070 8GB GPU. Both machines used Linux, Python version 3.6, PyTorch version 1.7.1 \cite{NEURIPS2019_9015}, and Scikit-learn.

\section{Results and Discussion}

\begin{figure*}[htb!]
\newlength{\plotwidth}
\setlength{\plotwidth}{.4\textwidth}
    \begin{subfigure}[b]{.4\textwidth}
        \begin{tikzpicture}
            \begin{axis}[%
                /pgf/number format/.cd,
                width =1.00\plotwidth,
                height=.62\plotwidth,
                scale only axis,
                ylabel={mAP@R on base test set},
                axis background/.style={fill=white},
                axis x line*=bottom,
                axis y line*=left,
                ymin = 0, ymax = .6, 
                xlabel style = {font=\bfseries},
                ylabel style={font=\bfseries},
                legend cell align={left},
                legend style={anchor=north east,legend columns=1, draw=none}
            ]
                \addplot[color=Accent-A,solid,line width=.6pt,mark=*]
                    table[x=classes_10, y=normal_triplet_cifar10, col sep=comma]{%
                       BaseResults.txt};
                \addlegendentry{Triplet loss};
                \addplot[color=Accent-B,solid,line width=.6pt,mark=*]
                    table[x=classes_10, y=normal_cont_cifar10, col sep=comma]{%
                       BaseResults.txt};
                \addlegendentry{Contrastive loss};
                \addplot[color=Accent-C,solid,line width=.6pt,mark=*]
                    table[x=classes_10, y=normal_center_cifar10, col sep=comma]{%
                        BaseResults.txt};
                \addlegendentry{Center loss};
                \addplot[color=Accent-E,solid,line width=.6pt,mark=*]
                    table[x=classes_10, y=normal_angular_cifar10, col sep=comma]{%
                        BaseResults.txt};
                \addlegendentry{Angular loss};
            \end{axis}
            \node[black,above right] at (.0,.2) {CIFAR10};
        \end{tikzpicture}%
    \end{subfigure} 
    \hspace{.1\textwidth}
    \begin{subfigure}[b]{.4\textwidth}%
        \begin{tikzpicture}
            \begin{axis}[%
                /pgf/number format/.cd,
                width =1.00\plotwidth,
                height=.62\plotwidth,
                scale only axis,
                label style={font=\bfseries},
                axis background/.style={fill=white},
                axis x line*=bottom,
                axis y line*=left,
                ymin = 0, ymax = 1, 
                xlabel style = {font=\bfseries},
                ylabel style={font=\bfseries}
            ]
                \addplot[color=Accent-A,solid,line width=.6pt,mark=*]
                    table[x=classes_10, y=normal_triplet_FMNIST, col sep=comma]{%
                        BaseResults.txt};
                \addplot[color=Accent-B,solid,line width=.6pt,mark=*]
                    table[x=classes_10, y=normal_cont_FMNIST, col sep=comma]{%
                        BaseResults.txt};
                \addplot[color=Accent-C,solid,line width=.6pt,mark=*]
                    table[x=classes_10, y=normal_center_FMNIST, col sep=comma]{%
                        BaseResults.txt};
                \addplot[color=Accent-E,solid,line width=.6pt,mark=*]
                    table[x=classes_10, y=normal_angular_FMNIST, col sep=comma]{%
                        BaseResults.txt};
            \end{axis}%
            \node[black,above right] at (.0,.2) {Fashion-MNIST};
        \end{tikzpicture}%
      \end{subfigure}  
      
    \begin{subfigure}[b]{.4\textwidth}%
        \begin{tikzpicture}
            \begin{axis}[%
                /pgf/number format/.cd,
                width =1.00\plotwidth,
                height=.62\plotwidth,
                scale only axis,
                xlabel={Number of classes incrementally trained},
                ylabel={mAP@R on base test set},
                axis background/.style={fill=white},
                axis x line*=bottom,
                axis y line*=left,
                ymin = 0, ymax = 1, 
                xlabel style = {font=\bfseries},
                ylabel style={font=\bfseries}
            ]
                \addplot[color=Accent-A,solid,line width=.6pt,mark=*]
                    table[x=classes_10, y=normal_triplet_MNIST, col sep=comma]{%
                        BaseResults.txt};
                \addplot[color=Accent-B,solid,line width=.6pt,mark=*]
                    table[x=classes_10, y=normal_cont_MNIST, col sep=comma]{%
                       BaseResults.txt};
                \addplot[color=Accent-C,solid,line width=.6pt,mark=*]
                    table[x=classes_10, y=normal_center_MNIST, col sep=comma]{%
                        BaseResults.txt};
                \addplot[color=Accent-E,solid,line width=.6pt,mark=*]
                    table[x=classes_10, y=normal_angular_MNIST, col sep=comma]{%
                    BaseResults.txt};
            \end{axis}%
            \node[black,above right] at (.0,.2) {MNIST};
        \end{tikzpicture}%
    \end{subfigure}  
    \hspace{.1\textwidth}
    \begin{subfigure}[b]{.4\textwidth}%
        \begin{tikzpicture}
            \begin{axis}[%
                /pgf/number format/.cd,
                width =1.00\plotwidth,
                height=.62\plotwidth,
                scale only axis,
                ymin = 0, ymax = 1, 
                xlabel={Number of classes incrementally trained},
                axis background/.style={fill=white},
                axis x line*=bottom,
                axis y line*=left,
                xlabel style = {font=\bfseries},
                ylabel style={font=\bfseries}
            ]
                \addplot[color=Accent-A,solid,line width=.6pt,mark=*]
                    table[x=classes_26, y=normal_triplet_EMNIST, col sep=comma]{%
                        BaseResults.txt};
                \addplot[color=Accent-B,solid,line width=.6pt,mark=*]
                    table[x=classes_26, y=normal_cont_EMNIST, col sep=comma]{%
                       BaseResults.txt};
                \addplot[color=Accent-C,solid,line width=.6pt,mark=*]
                    table[x=classes_26, y=normal_center_EMNIST, col sep=comma]{%
                       BaseResults.txt};
                \addplot[color=Accent-E,solid,line width=.6pt,mark=*]
                    table[x=classes_26, y=normal_angular_EMNIST, col sep=comma]{%
                    BaseResults.txt};
            \end{axis}%
            \node[black,above right] at (.0,.2) {EMNIST};
        \end{tikzpicture}%
      \end{subfigure} 
      \caption{Mean average precision on base-test set without using incremental learning techniques.}
      \label{fig:baseresultsnormal}
    \end{figure*}
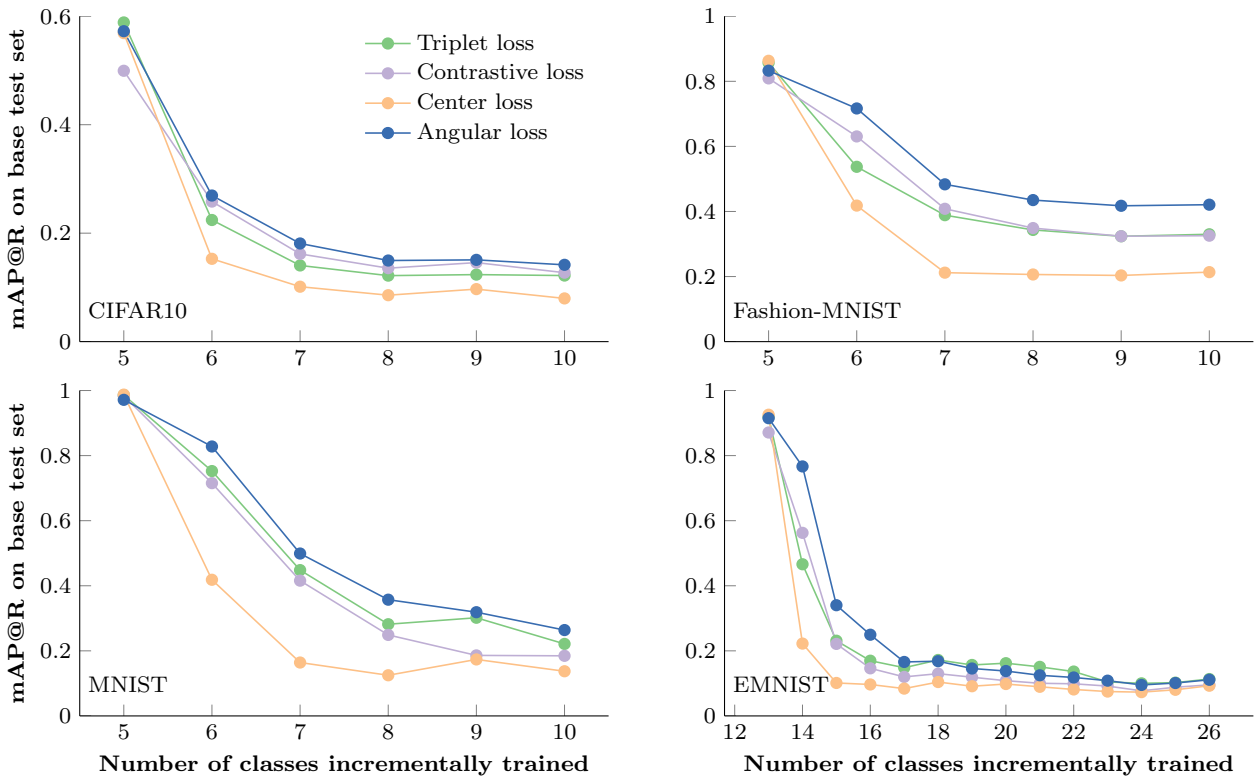 
The test metrics we used are a minor variation to the incremental class evaluation metrics of Kemker \emph{\textit{et al.}} \cite{kemker2018measuring} to support learning based on distance rather than classification. We measured performance (mAP@R) on the base-test set after learning classes sequentially. We use mAP@R since, in similarity learning, the loss functions we use do not make use of a classification layer, but instead, the networks are used to retrieve all related images of a query. The mAP@R metric is considered a superior metric for similarity learning in this regard~\cite{musgrave2020metric}. We tracked the model's performance on the new class to ensure it was still learning. We measured how well a model retains prior knowledge and learns new knowledge by measuring the mean mAP@R performance on each class learned during each training session. We used mAP shown in Equation \ref{eqn:MAP} to measure the performance of our model instead of the classification accuracy metric, as our models learn based on distance.

The evaluation metrics used are modified, to include mAP@R, versions of the original metrics by Kemker \emph{\textit{et al.}}\cite{kemker2018measuring} defined as:
\begin{equation}\label{eqn:Metric}
    \begin{aligned}
   \Omega_{base} &=  \frac{1}{T-1}\sum_{i=2}^{T}\frac{\alpha_{base,i}}{\alpha_{ideal}}\\
   \Omega_{new} &=  \frac{1}{T-1}\sum_{i=2}^{T} \alpha_{new,i}\\
   \Omega_{all} &=  \frac{1}{T-1}\sum_{i=2}^{T}\frac{\alpha_{all,i}}{\alpha_{ideal}}
   \end{aligned}
\end{equation}
where $T$ is the total number of training sessions, $\alpha_{new,i}$ is the test mAP@R for session $i$ immediately after it is learned. $\alpha_{base,i}$ is the test mAP@R on the ﬁrst session (base-test set) after i\textsuperscript{th} new sessions have been learned. $\alpha_{all,i}$ is the test mAP@R of the inc-test data and base-test set for the classes seen so far. $\alpha_{ideal}$ is the ofﬂine model mAP on the base-test set, which is the ideal performance. $\Omega_{base}$ measures a model’s retention of the base knowledge after sequential training sessions. $\Omega_{new}$ measures the model’s performance on new classes. $\Omega_{all}$ indicates how well a model retains prior knowledge and acquires new information (how well we retrieve newly learnt classes among previously seen classes). $\Omega_{base}$ and $\Omega_{all}$ are normalized with $\alpha_{ideal}$. The evaluation metrics are between [0,1] unless the results exceed the offline model. The offline models are trained on all the data.

The mAP@R is evaluated by extracting a feature embedding of size $128$ for every image. The images' embeddings are compared to all other embeddings and ranked in order of cosine similarity. Each test image was treated as a query once while comparing its similarity to the remaining images.

\begin{figure*}[htb!]
\setlength{\plotwidth}{.4\textwidth}
    \begin{subfigure}[b]{.4\textwidth}
        \begin{tikzpicture}
        \begin{axis}[%
          /pgf/number format/.cd,
          width =1.00\plotwidth,
          height=.62\plotwidth,
          scale only axis,
          ylabel={mAP@R on base test set},
          axis background/.style={fill=white},
          axis x line*=bottom,
          axis y line*=left,
          ymin = 0, ymax = .6, 
          xlabel style = {font=\bfseries},
          ylabel style={font=\bfseries},
          legend cell align={left},
          legend style={anchor=north east,legend columns=1, draw=none}
        ]
        \addplot[color=Dark2-A,solid,line width=1.0pt,mark=*]
        table[x=classes_10, y=ours_angular_cifar10, col sep=comma]{%
            Angular_Results.txt};
          \addlegendentry{Ours};
         \addplot[color=Dark2-B,solid,line width=.6pt,mark=*]
        table[x=classes_10, y=icarl_angular_cifar10, col sep=comma]{%
            Angular_Results.txt};
          \addlegendentry{iCarl};
         \addplot[color=Dark2-C,solid,line width=.6pt,mark=*]
        table[x=classes_10, y=ebll_angular_cifar10, col sep=comma]{%
            Angular_Results.txt};
          \addlegendentry{EBLL};
         \addplot[color=Dark2-D,solid,line width=.6pt,mark=*]
        table[x=classes_10, y=ewc_angular_cifar10, col sep=comma]{%
            Angular_Results.txt};
          \addlegendentry{EWC};
         \addplot[color=Dark2-F,solid,line width=.6pt,mark=*]
        table[x=classes_10, y=normal_angular_cifar10, col sep=comma]{%
            Angular_Results.txt};
        \addplot[color=Dark2-H,solid,dashed,width=1.0pt,mark=o]
        table[x=classes_10, y=ideal_angular_cifar10, col sep=comma]{%
            Angular_Results.txt};
          \addlegendentry{Normal};
        \end{axis}
        \node[black,above right] at (.0,.2) {CIFAR10};
        \end{tikzpicture}%
      \end{subfigure}
      \hspace{.1\textwidth}
      \begin{subfigure}[b]{.4\textwidth}%
        \begin{tikzpicture}
          \begin{axis}[%
          /pgf/number format/.cd,
          width =1.00\plotwidth,
          height=.62\plotwidth,
          scale only axis,
          axis background/.style={fill=white},
          axis x line*=bottom,
          axis y line*=left,
          ymin = 0,
          ymax = 1,
          xlabel style={font=\bfseries},
          ylabel style={font=\bfseries}
          ]
         \addplot[color=Dark2-A,solid,line width=1.0pt,mark=*]
        table[x=classes_10, y=ours_angular_FMNIST, col sep=comma]{%
           Angular_Results.txt};
         \addplot[color=Dark2-B,solid,line width=.6pt,mark=*]
        table[x=classes_10, y=icarl_angular_FMNIST, col sep=comma]{%
           Angular_Results.txt};

         \addplot[color=Dark2-C,solid,line width=.6pt,mark=*]
        table[x=classes_10, y=ebll_angular_FMNIST, col sep=comma]{%
            Angular_Results.txt};

         \addplot[color=Dark2-D,solid,line width=.6pt,mark=*]
        table[x=classes_10, y=ewc_angular_FMNIST, col sep=comma]{%
            Angular_Results.txt};

         \addplot[color=Dark2-F,solid,line width=.6pt,mark=*]
        table[x=classes_10, y=normal_angular_FMNIST, col sep=comma]{%
            Angular_Results.txt};
        \addplot[color=Dark2-H,solid,dashed,width=1.0pt,mark=o]
        table[x=classes_10, y=ideal_angular_FMNIST, col sep=comma]{%
           Angular_Results.txt};

          \end{axis}%
          \node[black,above right] at (.0,.2) {Fashion-MNIST};
        \end{tikzpicture}%
      \end{subfigure}  
      
      \begin{subfigure}[b]{.4\textwidth}%
        \begin{tikzpicture}
          \begin{axis}[%
          /pgf/number format/.cd,
          width =1.00\plotwidth,
          height=.62\plotwidth,
          scale only axis,
          xlabel={Number of classes incrementally trained},
          ylabel={mAP@R on base test set},
          axis background/.style={fill=white},
          axis x line*=bottom,
          axis y line*=left,
          ymin = 0,
          ymax = 1,
          xlabel style={font=\bfseries},
          ylabel style={font=\bfseries}
          ]
         \addplot[color=Dark2-A,solid,line width=1.0pt,mark=*]
        table[x=classes_10, y=ours_angular_MNIST, col sep=comma]{%
            Angular_Results.txt};

         \addplot[color=Dark2-B,solid,line width=.6pt,mark=*]
        table[x=classes_10, y=icarl_angular_MNIST, col sep=comma]{%
            Angular_Results.txt};
  
         \addplot[color=Dark2-C,solid,line width=.6pt,mark=*]
        table[x=classes_10, y=ebll_angular_MNIST, col sep=comma]{%
           Angular_Results.txt};

         \addplot[color=Dark2-D,solid,line width=.6pt,mark=*]
        table[x=classes_10, y=ewc_angular_MNIST, col sep=comma]{%
            Angular_Results.txt};

         \addplot[color=Dark2-F,solid,line width=.6pt,mark=*]
        table[x=classes_10, y=normal_angular_MNIST, col sep=comma]{%
            Angular_Results.txt};
        \addplot[color=Dark2-H,solid,dashed,width=1.0pt,mark=o]
        table[x=classes_10, y=ideal_angular_MNIST, col sep=comma]{%
            Angular_Results.txt};
          \end{axis}%
          \node[black,above right] at (.0,.2) {MNIST};
        \end{tikzpicture}%
      \end{subfigure}  
      \hspace{.1\textwidth}
      \begin{subfigure}[b]{.4\textwidth}%
        \begin{tikzpicture}
          \begin{axis}[%
          /pgf/number format/.cd,
          width =1.00\plotwidth,
          height=.62\plotwidth,
          scale only axis,
          xlabel={Number of classes incrementally trained},
          axis background/.style={fill=white},
          axis x line*=bottom,
          axis y line*=left,
          ymin = 0,
          ymax = 1,
          xlabel style= {font=\bfseries},
          ylabel style={font=\bfseries}
          ]
         \addplot[color=Dark2-A,solid,line width=1.0pt,mark=*]
        table[x=classes_26, y=ours_angular_EMNIST, col sep=comma]{%
            Angular_Results.txt};

         \addplot[color=Dark2-B,solid,line width=.6pt,mark=*]
        table[x=classes_26, y=icarl_angular_EMNIST, col sep=comma]{%
           Angular_Results.txt};
  
         \addplot[color=Dark2-C,solid,line width=.6pt,mark=*]
        table[x=classes_26, y=ebll_angular_EMNIST, col sep=comma]{%
           Angular_Results.txt};

         \addplot[color=Dark2-D,solid,line width=.6pt,mark=*]
        table[x=classes_26, y=ewc_angular_EMNIST, col sep=comma]{%
            Angular_Results.txt};

         \addplot[color=Dark2-F,solid,line width=.6pt,mark=*]
        table[x=classes_26, y=normal_angular_EMNIST, col sep=comma]{%
            Angular_Results.txt};
        \addplot[color=Dark2-H,solid,dashed,width=1.0pt,mark=o]
        table[x=classes_26, y=ideal_angular_EMNIST, col sep=comma]{%
            Angular_Results.txt};
            
          \end{axis}%
          \node[black,above right] at (.0,.2) {EMNIST};
        \end{tikzpicture}%
      \end{subfigure} 
      \caption{The figure compares the different incremental learning approaches to normal training for \textbf{Angular loss} as a baseline. Mean average precision (mAP@R) on base-test set. Solid lines indicate incremental learning models. The dotted line indicates the offline ideal.}
      \label{fig:angularresults}
    \end{figure*}
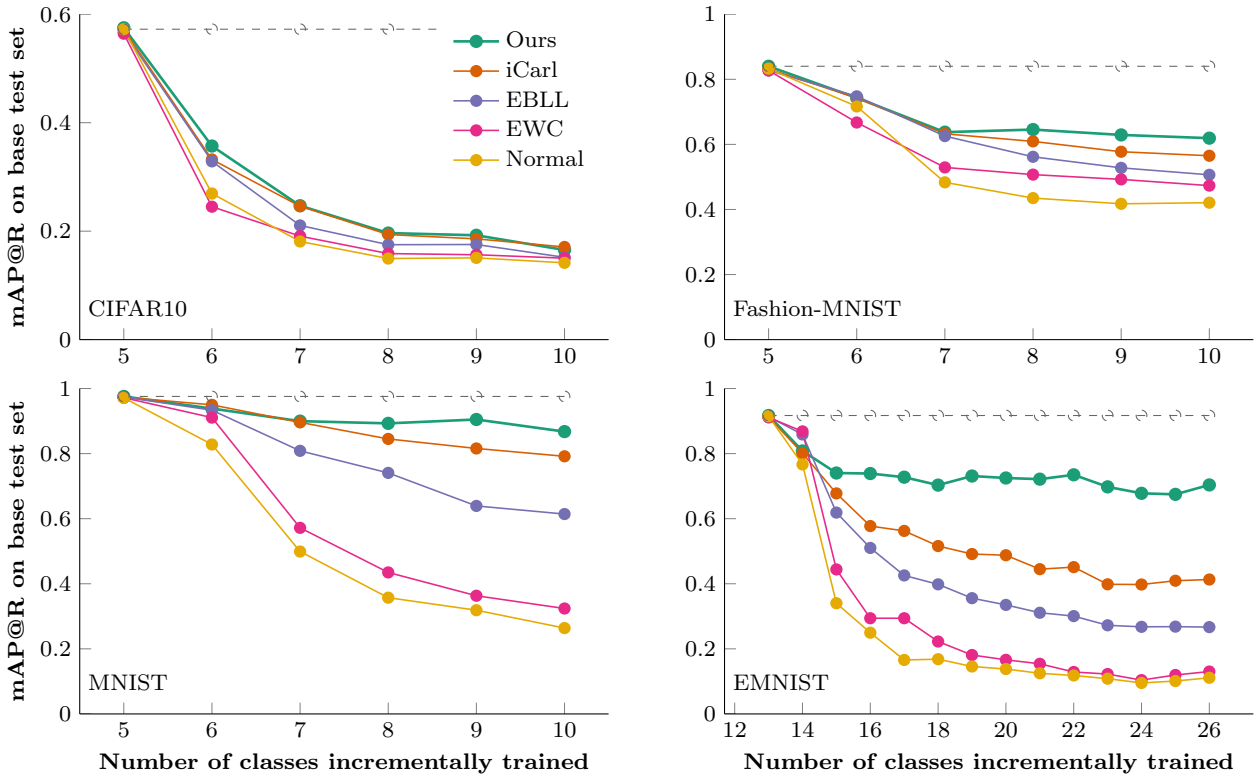

\begin{figure*}[htb!]
      \setlength{\plotwidth}{.4\textwidth}
      \begin{subfigure}[b]{.4\textwidth}
        \begin{tikzpicture}
          \begin{axis}[%
          /pgf/number format/.cd,
          width =1.00\plotwidth,
          height=.62\plotwidth,
          scale only axis,
          ylabel={mAP@R on base test set},
          axis background/.style={fill=white},
          axis x line*=bottom,
          axis y line*=left,
          ymin = 0, ymax = .6, 
          xlabel style = {font=\bfseries},
          ylabel style={font=\bfseries},
          legend cell align={left},
          legend style={anchor=north east,legend columns=1, draw=none}
          ]
         \addplot[color=Dark2-A,solid,line width=1.0pt,mark=*]
        table[x=classes_10, y=ours_triplet_cifar10, col sep=comma]{%
            TripletResults.txt};
          \addlegendentry{Ours};
         \addplot[color=Dark2-B,solid,line width=.6pt,mark=*]
        table[x=classes_10, y=icarl_triplet_cifar10, col sep=comma]{%
            TripletResults.txt};
          \addlegendentry{iCarl};
         \addplot[color=Dark2-C,solid,line width=.6pt,mark=*]
        table[x=classes_10, y=ebll_triplet_cifar10, col sep=comma]{%
            TripletResults.txt};
          \addlegendentry{EBLL};
         \addplot[color=Dark2-D,solid,line width=.6pt,mark=*]
        table[x=classes_10, y=ewc_triplet_cifar10, col sep=comma]{%
            TripletResults.txt};
          \addlegendentry{EWC};
         \addplot[color=Dark2-F,solid,line width=.6pt,mark=*]
        table[x=classes_10, y=normal_triplet_cifar10, col sep=comma]{%
            TripletResults.txt};
          \addlegendentry{Normal};
         \addplot[color=Dark2-H,solid,dashed,width=1.0pt,mark=o]
        table[x=classes_10, y=ideal_triplet_cifar10, col sep=comma]{%
            TripletResults.txt};
          \end{axis}
          \node[black,above right] at (.0,.2) {CIFAR10};
        \end{tikzpicture}%
      \end{subfigure} 
       \hspace{.1\textwidth}
      \begin{subfigure}[b]{.4\textwidth}%
        \begin{tikzpicture}
          \begin{axis}[%
          /pgf/number format/.cd,
          width =1.00\plotwidth,
          height=.62\plotwidth,
          scale only axis,
          axis background/.style={fill=white},
          axis x line*=bottom,
          axis y line*=left,
          ymin = 0, ymax = 1, 
          xlabel style = {font=\bfseries},
          ylabel style={font=\bfseries}
          ]
         \addplot[color=Dark2-A,solid,line width=1.0pt,mark=*]
        table[x=classes_10, y=ours_triplet_FMNIST, col sep=comma]{%
            TripletResults.txt};

         \addplot[color=Dark2-B,solid,line width=.6pt,mark=*]
        table[x=classes_10, y=icarl_triplet_FMNIST, col sep=comma]{%
            TripletResults.txt};

         \addplot[color=Dark2-C,solid,line width=.6pt,mark=*]
        table[x=classes_10, y=ebll_triplet_FMNIST, col sep=comma]{%
            TripletResults.txt};

         \addplot[color=Dark2-D,solid,line width=.6pt,mark=*]
        table[x=classes_10, y=ewc_triplet_FMNIST, col sep=comma]{%
            TripletResults.txt};

         \addplot[color=Dark2-F,solid,line width=.6pt,mark=*]
        table[x=classes_10, y=normal_triplet_FMNIST, col sep=comma]{%
            TripletResults.txt};
        \addplot[color=Dark2-H,solid,dashed,width=1.0pt,mark=o]
        table[x=classes_10, y=ideal_triplet_FMNIST, col sep=comma]{%
            TripletResults.txt};
          
          \end{axis}%
          \node[black,above right] at (.0,.2) {Fashion-MNIST};
        \end{tikzpicture}%
      \end{subfigure}  
      
      \begin{subfigure}[b]{.4\textwidth}%
        \begin{tikzpicture}
          \begin{axis}[%
          /pgf/number format/.cd,
          width =1.00\plotwidth,
          height=.62\plotwidth,
          scale only axis,
          ylabel={mAP@R on base test set},
          xlabel={Number of classes incrementally trained},
          axis background/.style={fill=white},
          axis x line*=bottom,
          axis y line*=left,
          ymin = 0, ymax = 1, 
          xlabel style = {font=\bfseries},
          ylabel style={font=\bfseries}
          ]
         \addplot[color=Dark2-A,solid,line width=1.0pt,mark=*]
        table[x=classes_10, y=ours_triplet_MNIST, col sep=comma]{%
            TripletResults.txt};

         \addplot[color=Dark2-B,solid,line width=.6pt,mark=*]
        table[x=classes_10, y=icarl_triplet_MNIST, col sep=comma]{%
            TripletResults.txt};

         \addplot[color=Dark2-C,solid,line width=.6pt,mark=*]
        table[x=classes_10, y=ebll_triplet_MNIST, col sep=comma]{%
            TripletResults.txt};

         \addplot[color=Dark2-D,solid,line width=.6pt,mark=*]
        table[x=classes_10, y=ewc_triplet_MNIST, col sep=comma]{%
            TripletResults.txt};

         \addplot[color=Dark2-F,solid,line width=.6pt,mark=*]
        table[x=classes_10, y=normal_triplet_MNIST, col sep=comma]{%
            TripletResults.txt};
        \addplot[color=Dark2-H,solid,dashed,width=1.0pt,mark=o]
        table[x=classes_10, y=ideal_triplet_MNIST, col sep=comma]{%
            TripletResults.txt};
          \end{axis}%
          \node[black,above right] at (.0,.2) {MNIST};
        \end{tikzpicture}%
      \end{subfigure}  
      \hspace{.1\textwidth}
      \begin{subfigure}[b]{.4\textwidth}%
        \begin{tikzpicture}
          \begin{axis}[%
          /pgf/number format/.cd,
          width =1.00\plotwidth,
          height=.62\plotwidth,
          scale only axis,
          xlabel={Number of classes incrementally trained},
          axis background/.style={fill=white},
          axis x line*=bottom,
          axis y line*=left,
          ymin = 0, ymax = 1, 
          xlabel style = {font=\bfseries},
          ylabel style={font=\bfseries}
          ]
         \addplot[color=Dark2-A,solid,line width=1.0pt,mark=*]
        table[x=classes_26, y=ours_triplet_EMNIST, col sep=comma]{%
            TripletResults.txt};

         \addplot[color=Dark2-B,solid,line width=.6pt,mark=*]
        table[x=classes_26, y=icarl_triplet_EMNIST, col sep=comma]{%
            TripletResults.txt};

         \addplot[color=Dark2-C,solid,line width=.6pt,mark=*]
        table[x=classes_26, y=ebll_triplet_EMNIST, col sep=comma]{%
            TripletResults.txt};

         \addplot[color=Dark2-D,solid,line width=.6pt,mark=*]
        table[x=classes_26, y=ewc_triplet_EMNIST, col sep=comma]{%
            TripletResults.txt};

         \addplot[color=Dark2-F,solid,line width=.6pt,mark=*]
        table[x=classes_26, y=normal_triplet_EMNIST, col sep=comma]{%
            TripletResults.txt};
        \addplot[color=Dark2-H,solid,dashed,width=1.0pt,mark=o]
        table[x=classes_26, y=ideal_triplet_EMNIST, col sep=comma]{%
            TripletResults.txt};
          \end{axis}%
            \node[black,above right] at (.0,.2) {EMNIST};
        \end{tikzpicture}%
      \end{subfigure} 
      \caption{The figure compares the different incremental learning approaches to normal training for \textbf{Triplet loss} as a baseline. Mean average precision (mAP@R) on base-test set. Solid lines indicate incremental learning models. The dotted line indicates the offline ideal. }
      \label{fig:tripletresults}
    \end{figure*}
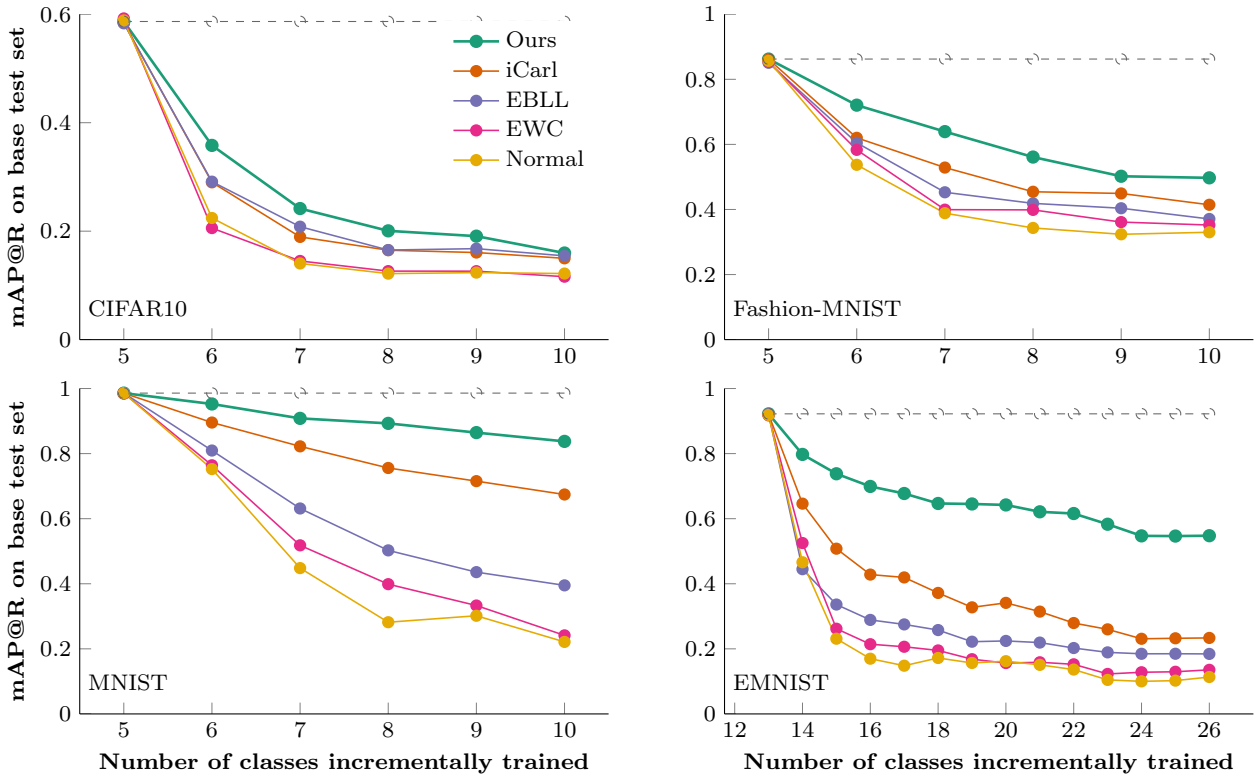
    
    
\begin{figure*}[htb!]
      \setlength{\plotwidth}{.4\textwidth}
      \begin{subfigure}[b]{.4\textwidth}
        \begin{tikzpicture}
          \begin{axis}[%
          /pgf/number format/.cd,
          width =1.00\plotwidth,
          height=.62\plotwidth,
          scale only axis,
          ylabel={mAP@R on base test set},
          axis background/.style={fill=white},
          axis x line*=bottom,
          axis y line*=left,
          ymin = 0, ymax = .6, 
          xlabel style = {font=\bfseries},
          ylabel style={font=\bfseries},
          legend cell align={left},
          legend style={anchor=north east,legend columns=1, draw=none}
          ]
         \addplot[color=Dark2-A,solid,line width=1.0pt,mark=*]        table[x=classes_10, y=ours_cont_cifar10, col sep=comma]{%
            ContrastiveResults.txt};
          \addlegendentry{Ours};
         \addplot[color=Dark2-B,solid,line width=.6pt,mark=*]
        table[x=classes_10, y=icarl_cont_cifar10, col sep=comma]{%
            ContrastiveResults.txt};
          \addlegendentry{iCaRL};
         \addplot[color=Dark2-C,solid,line width=.6pt,mark=*]
        table[x=classes_10, y=ebll_cont_cifar10, col sep=comma]{%
            ContrastiveResults.txt};
          \addlegendentry{EBLL};
         \addplot[color=Dark2-D,solid,line width=.6pt,mark=*]
        table[x=classes_10, y=ewc_cont_cifar10, col sep=comma]{%
            ContrastiveResults.txt};
          \addlegendentry{EWC};
         \addplot[color=Dark2-F,solid,line width=.6pt,mark=*]
        table[x=classes_10, y=normal_cont_cifar10, col sep=comma]{%
            ContrastiveResults.txt};
          \addlegendentry{Normal};
        \addplot[color=Dark2-H,solid,dashed,width=1.0pt,mark=o]
        table[x=classes_10, y=ideal_cont_cifar10, col sep=comma]{%
           ContrastiveResults.txt};
            
          \end{axis}
          \node[black,above right] at (.0,.2) {CIFAR10};
        \end{tikzpicture}%
      \end{subfigure} 
      \hspace{.1\textwidth}
      \begin{subfigure}[b]{.4\textwidth}%
        \begin{tikzpicture}
          \begin{axis}[%
          /pgf/number format/.cd,
          width =1.00\plotwidth,
          height=.62\plotwidth,
          scale only axis,
          axis background/.style={fill=white},
          axis x line*=bottom,
          axis y line*=left,
          ymin = 0, ymax = 1, 
          xlabel style={font=\bfseries},
          ylabel style={font=\bfseries}
          ]
         \addplot[color=Dark2-A,solid,line width=1.0pt,mark=*]
        table[x=classes_10, y=ours_cont_FMNIST, col sep=comma]{%
           ContrastiveResults.txt};
         \addplot[color=Dark2-B,solid,line width=.6pt,mark=*]
        table[x=classes_10, y=icarl_cont_FMNIST, col sep=comma]{%
            ContrastiveResults.txt};
  
         \addplot[color=Dark2-C,solid,line width=.6pt,mark=*]
        table[x=classes_10, y=ebll_cont_FMNIST, col sep=comma]{%
            ContrastiveResults.txt};
  
         \addplot[color=Dark2-D,solid,line width=.6pt,mark=*]
        table[x=classes_10, y=ewc_cont_FMNIST, col sep=comma]{%
            ContrastiveResults.txt};

         \addplot[color=Dark2-F,solid,line width=.6pt,mark=*]
        table[x=classes_10, y=normal_cont_FMNIST, col sep=comma]{%
            ContrastiveResults.txt};
        \addplot[color=Dark2-H,solid,dashed,width=1.0pt,mark=o]
        table[x=classes_10, y=ideal_cont_FMNIST, col sep=comma]{%
            ContrastiveResults.txt};
          \end{axis}%
          \node[black,above right] at (.0,.2) {Fashion-MNIST};
        \end{tikzpicture}%
      \end{subfigure}  
      
      \begin{subfigure}[b]{.4\textwidth}%
        \begin{tikzpicture}
          \begin{axis}[%
          /pgf/number format/.cd,
          width =1.00\plotwidth,
          height=.62\plotwidth,
          scale only axis,
          ylabel={mAP@R on base test set},
          xlabel={Number of classes incrementally trained},
          axis background/.style={fill=white},
          axis x line*=bottom,
          axis y line*=left,
          ymin = 0, ymax = 1, 
          xlabel style={font=\bfseries},
          ylabel style={font=\bfseries}
          ]
         \addplot[color=Dark2-A,solid,line width=1.0pt,mark=*]
        table[x=classes_10, y=ours_cont_MNIST, col sep=comma]{%
            ContrastiveResults.txt};
         \addplot[color=Dark2-B,solid,line width=.6pt,mark=*]
        table[x=classes_10, y=icarl_cont_MNIST, col sep=comma]{%
            ContrastiveResults.txt};
  
         \addplot[color=Dark2-C,solid,line width=.6pt,mark=*]
        table[x=classes_10, y=ebll_cont_MNIST, col sep=comma]{%
            ContrastiveResults.txt};
  
         \addplot[color=Dark2-D,solid,line width=.6pt,mark=*]
        table[x=classes_10, y=ewc_cont_MNIST, col sep=comma]{%
            ContrastiveResults.txt};

         \addplot[color=Dark2-F,solid,line width=.6pt,mark=*]
        table[x=classes_10, y=normal_cont_MNIST, col sep=comma]{%
            ContrastiveResults.txt};
        \addplot[color=Dark2-H,solid,dashed,width=1.0pt,mark=o]
        table[x=classes_10, y=ideal_cont_MNIST, col sep=comma]{%
            ContrastiveResults.txt};
          \end{axis}%
          \node[black,above right] at (.0,.2) {MNIST};
        \end{tikzpicture}%
      \end{subfigure}  
      \hspace{.1\textwidth}
      \begin{subfigure}[b]{.4\textwidth}%
        \begin{tikzpicture}
          \begin{axis}[%
          /pgf/number format/.cd,
          width =1.00\plotwidth,
          height=.62\plotwidth,
          scale only axis,
          xlabel={Number of classes incrementally trained},
          axis background/.style={fill=white},
          axis x line*=bottom,
          axis y line*=left,
          ymin = 0, ymax = 1, 
          xlabel style = {font=\bfseries},
          ylabel style={font=\bfseries}
          ]
          \addplot[color=Dark2-A,solid,line width=1.0pt,mark=*]
        table[x=classes_26, y=ours_cont_EMNIST, col sep=comma]{%
            ContrastiveResults.txt};
         \addplot[color=Dark2-B,solid,line width=.6pt,mark=*]
        table[x=classes_26, y=icarl_cont_EMNIST, col sep=comma]{%
            ContrastiveResults.txt};
         \addplot[color=Dark2-C,solid,line width=.6pt,mark=*]
        table[x=classes_26, y=ebll_cont_EMNIST, col sep=comma]{%
            ContrastiveResults.txt};
         \addplot[color=Dark2-D,solid,line width=.6pt,mark=*]
        table[x=classes_26, y=ewc_cont_EMNIST, col sep=comma]{%
            ContrastiveResults.txt};
         \addplot[color=Dark2-F,solid,line width=.6pt,mark=*]
        table[x=classes_26, y=normal_cont_EMNIST, col sep=comma]{%
            ContrastiveResults.txt};
        \addplot[color=Dark2-H,solid,dashed,width=1.0pt,mark=o]
        table[x=classes_26, y=ideal_cont_EMNIST, col sep=comma]{%
           ContrastiveResults.txt};
          \end{axis}%
        \node[black,above right] at (.0,.2) {EMNIST};
        \end{tikzpicture}%
      \end{subfigure} 
      \caption{The figure compares the different incremental learning approaches to normal training for \textbf{Contrastive loss} as a baseline. Mean average precision (mAP@R) on base-test set. Solid lines indicate incremental learning models. The dotted line indicates the offline ideal. }
      \label{fig:contrastiveresults}
    \end{figure*}
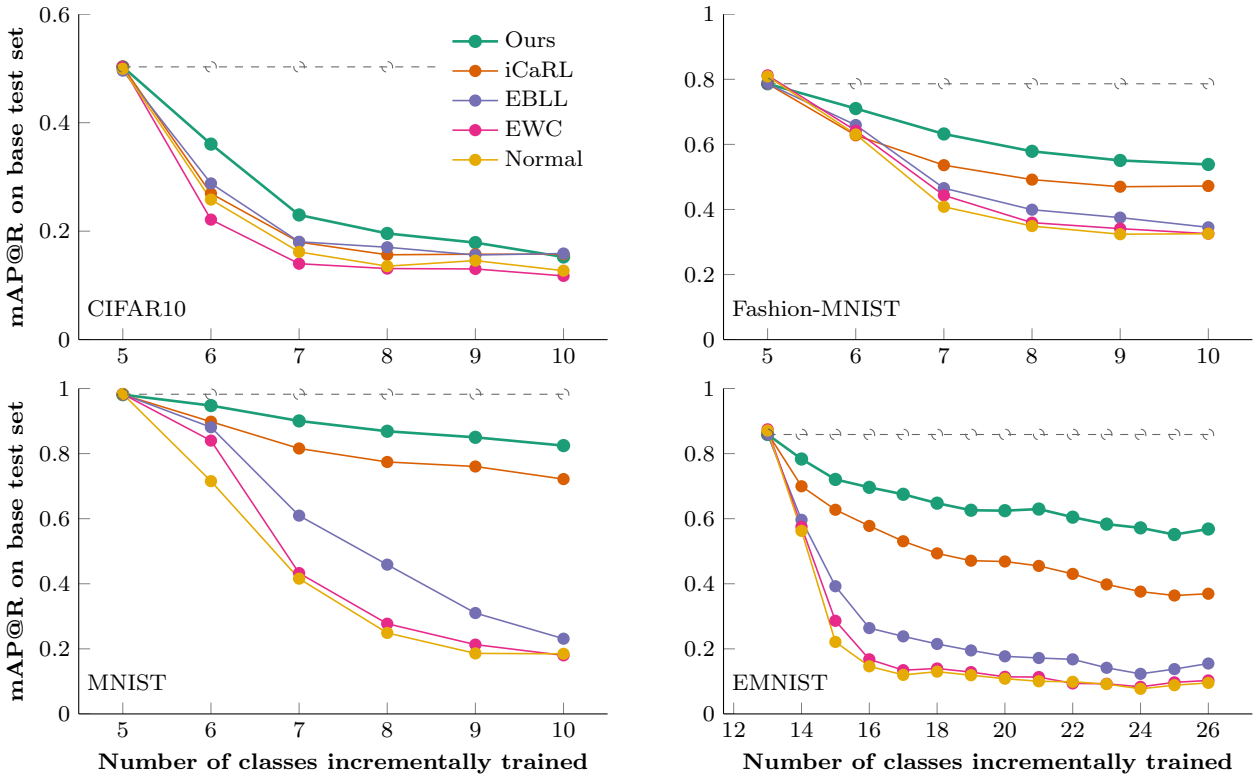
    
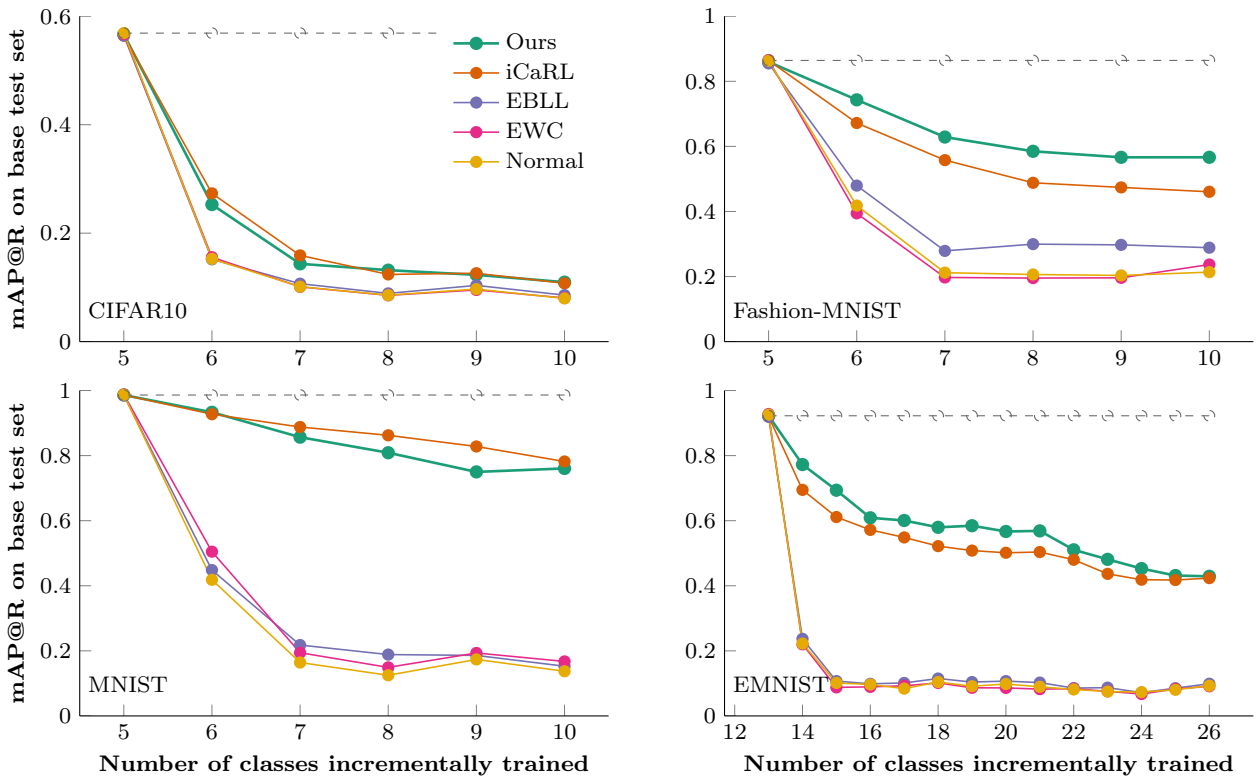
\begin{figure*}[htb!]
      \setlength{\plotwidth}{.4\textwidth}
      \begin{subfigure}[b]{.4\textwidth}
        \begin{tikzpicture}
          \begin{axis}[%
          /pgf/number format/.cd,
          width =1.00\plotwidth,
          height=.62\plotwidth,
          scale only axis,
          ylabel={mAP@R on base test set},
          axis background/.style={fill=white},
          axis x line*=bottom,
          axis y line*=left,
          ymin = 0, ymax = .6, 
          xlabel style = {font=\bfseries},
          ylabel style={font=\bfseries},
          legend cell align={left},
          legend style={anchor=north east,legend columns=1, draw=none}
          ]
         \addplot[color=Dark2-A,solid,line width=1.0pt,mark=*]
        table[x=classes_10, y=ours_center_cifar10, col sep=comma]{%
            CenterResults.txt};
          \addlegendentry{Ours};
         \addplot[color=Dark2-B,solid,line width=.6pt,mark=*]
        table[x=classes_10, y=icarl_center_cifar10, col sep=comma]{%
            CenterResults.txt};
          \addlegendentry{iCaRL};
         \addplot[color=Dark2-C,solid,line width=.6pt,mark=*]
        table[x=classes_10, y=ebll_center_cifar10, col sep=comma]{%
            CenterResults.txt};
          \addlegendentry{EBLL};
         \addplot[color=Dark2-D,solid,line width=.6pt,mark=*]
        table[x=classes_10, y=ewc_center_cifar10, col sep=comma]{%
            CenterResults.txt};
          \addlegendentry{EWC};
         \addplot[color=Dark2-F,solid,line width=.6pt,mark=*]
        table[x=classes_10, y=normal_center_cifar10, col sep=comma]{%
            CenterResults.txt};
          \addlegendentry{Normal};
        \addplot[color=Dark2-H,solid,dashed,width=1.0pt,mark=o]
        table[x=classes_10, y=ideal_center_cifar10, col sep=comma]{%
            CenterResults.txt};
          \end{axis}
          \node[black,above right] at (.0,.2) {CIFAR10};
        \end{tikzpicture}%
      \end{subfigure} 
       \hspace{.1\textwidth}
      \begin{subfigure}[b]{.4\textwidth}%
        \begin{tikzpicture}
          \begin{axis}[%
          /pgf/number format/.cd,
          width =1.00\plotwidth,
          height=.62\plotwidth,
          scale only axis,
          axis background/.style={fill=white},
          axis x line*=bottom,
          axis y line*=left,
          ymin = 0, ymax = 1, 
          xlabel style = {font=\bfseries},
          ylabel style={font=\bfseries}
          ]
         \addplot[color=Dark2-A,solid,line width=1.0pt,mark=*]
        table[x=classes_10, y=ours_center_FMNIST, col sep=comma]{%
            CenterResults.txt};
         \addplot[color=Dark2-B,solid,line width=.6pt,mark=*]
        table[x=classes_10, y=icarl_center_FMNIST, col sep=comma]{%
            CenterResults.txt};
         \addplot[color=Dark2-C,solid,line width=.6pt,mark=*]
        table[x=classes_10, y=ebll_center_FMNIST, col sep=comma]{%
            CenterResults.txt};
         \addplot[color=Dark2-D,solid,line width=.6pt,mark=*]
        table[x=classes_10, y=ewc_center_FMNIST, col sep=comma]{%
            CenterResults.txt};
         \addplot[color=Dark2-F,solid,line width=.6pt,mark=*]
        table[x=classes_10, y=normal_center_FMNIST, col sep=comma]{%
            CenterResults.txt};
        \addplot[color=Dark2-H,solid,dashed,width=1.0pt,mark=o]
        table[x=classes_10, y=ideal_center_FMNIST, col sep=comma]{%
            CenterResults.txt};
          \end{axis}%
          \node[black,above right] at (.0,.2) {Fashion-MNIST};
        \end{tikzpicture}%
      \end{subfigure}  
      
      \begin{subfigure}[b]{.4\textwidth}%
        \begin{tikzpicture}
          \begin{axis}[%
          /pgf/number format/.cd,
          width =1.00\plotwidth,
          height=.62\plotwidth,
          scale only axis,
          ylabel={mAP@R on base test set},
          xlabel={Number of classes incrementally trained},
          axis background/.style={fill=white},
          axis x line*=bottom,
          axis y line*=left,
          ymin = 0, ymax = 1, 
          xlabel style = {font=\bfseries},
          ylabel style={font=\bfseries}
          ]
         \addplot[color=Dark2-A,solid,line width=1.0pt,mark=*]
        table[x=classes_10, y=ours_center_MNIST, col sep=comma]{%
            CenterResults.txt};
         \addplot[color=Dark2-B,solid,line width=.6pt,mark=*]
        table[x=classes_10, y=icarl_center_MNIST, col sep=comma]{%
            CenterResults.txt};
         \addplot[color=Dark2-C,solid,line width=.6pt,mark=*]
        table[x=classes_10, y=ebll_center_MNIST, col sep=comma]{%
            CenterResults.txt};
         \addplot[color=Dark2-D,solid,line width=.6pt,mark=*]
        table[x=classes_10, y=ewc_center_MNIST, col sep=comma]{%
            CenterResults.txt};
         \addplot[color=Dark2-F,solid,line width=.6pt,mark=*]
        table[x=classes_10, y=normal_center_MNIST, col sep=comma]{%
            CenterResults.txt};
        \addplot[color=Dark2-H,solid,dashed,width=1.0pt,mark=o]
        table[x=classes_10, y=ideal_center_MNIST, col sep=comma]{%
            CenterResults.txt};
          \end{axis}%
          \node[black,above right] at (.0,.2) {MNIST};
        \end{tikzpicture}%
      \end{subfigure}  
      \hspace{.1\textwidth}
      \begin{subfigure}[b]{.4\textwidth}%
        \begin{tikzpicture}
          \begin{axis}[%
          /pgf/number format/.cd,
          width =1.00\plotwidth,
          height=.62\plotwidth,
          scale only axis,
          xlabel={Number of classes incrementally trained},
          axis background/.style={fill=white},
          axis x line*=bottom,
          axis y line*=left,
          ymin = 0, ymax = 1, 
          xlabel style = {font=\bfseries},
          ylabel style={font=\bfseries}
          ]
         \addplot[color=Dark2-A,solid,line width=1.0pt,mark=*]
        table[x=classes_26, y=ours_center_EMNIST, col sep=comma]{%
            CenterResults.txt};
         \addplot[color=Dark2-B,solid,line width=.6pt,mark=*]
        table[x=classes_26, y=icarl_center_EMNIST, col sep=comma]{%
            CenterResults.txt};
         \addplot[color=Dark2-C,solid,line width=.6pt,mark=*]
        table[x=classes_26, y=ebll_center_EMNIST, col sep=comma]{%
            CenterResults.txt};
         \addplot[color=Dark2-D,solid,line width=.6pt,mark=*]
        table[x=classes_26, y=ewc_center_EMNIST, col sep=comma]{%
            CenterResults.txt};
         \addplot[color=Dark2-F,solid,line width=.6pt,mark=*]
        table[x=classes_26, y=normal_center_EMNIST, col sep=comma]{%
            CenterResults.txt};
        \addplot[color=Dark2-H,solid,dashed,width=1.0pt,mark=o]
        table[x=classes_26, y=ideal_center_EMNIST, col sep=comma]{%
            CenterResults.txt};
          \end{axis}%
          \node[black,above right] at (.0,.2) {EMNIST};
        \end{tikzpicture}%
      \end{subfigure} 
      \caption{The figure compares the different incremental learning approaches to normal training for \textbf{Center loss} as a baseline. Mean average precision (mAP@R) on base-test set. Solid lines indicate incremental learning models. The dotted line indicates the offline ideal. }
      \label{fig:centerresults}
    \end{figure*}

\subsection{Test on Base Test Set During Sequential Learning}

Figure~\ref{fig:baseresultsnormal} shows how catastrophic forgetting affects initial base knowledge learned during incremental learning. The work by Huo \emph{\textit{et al.}} \cite{9311580} showed that Triplet loss is less affected by catastrophic forgetting when compared to other similarity learning functions. However, there were no special mining techniques for Contrastive and Angular loss in their work. In the current setup, we have introduced a pairwise and Angular miner for Contrastive and Angular loss. With the correct setup of Pair and Triplet mining for the similarity learning functions, the results show that the various methods suffer from catastrophic forgetting at approximately the same rate. Figure~\ref{fig:baseresultsnormal} shows that Center loss is still the most impacted by catastrophic forgetting compared to the other similarity learning loss functions.

Figures \ref{fig:angularresults} to \ref{fig:centerresults} highlight how each of the methods implemented reduces catastrophic forgetting during sequential class learning by testing on a base-test set after each new class is introduced. Offline models were trained on all available classes and tested on the base-test set to get the ideal mAP@R shown by a dotted line. Figure \ref{fig:angularresults} shows our method outperforms the baseline methods on FMNIST, MNIST, and EMNIST. There is no significant difference between iCaRL and our approach shown in the CIFAR10 result. Figure \ref{fig:tripletresults} and Figure \ref{fig:contrastiveresults} present our method as the best performer with Triplet and Contrastive loss across all the datasets. Figure \ref{fig:centerresults} shows iCaRL outperforming our method on CIFAR10 and MNIST but not on FMNIST and EMNIST. It is unclear whether iCaRL or our method performs best with Center loss.

The EMNIST results highlight the complete forgetting of base knowledge if there has been no effort to preserve knowledge over a long time and reinforce this research's motivation. We note a steeper drop in performance dependent on the number of classes previously learned, as shown for the EMNIST dataset. The complexity of the dataset equally affects the drop's steepness, as seen in the CIFAR10 case. We see earlier and more severe catastrophic forgetting in more complex datasets.

The EMNIST results in Figure~\ref{fig:angularresults} also illustrate some of the challenges associated with the iCaRL method's retained exemplars. As the number of exemplars from previously seen classes decreases, the network suffers from increased catastrophic forgetting. Secondly, variations of images are essential and keeping only exemplars closest to the class's mean does not represent a class well. The effectiveness of the exemplars seems to depend on how the mining and loss function work together. The figures show differences in forgetting rates between the loss functions and the same number of exemplars.

In Figure \ref{fig:angularresults} to \ref{fig:centerresults}, we observe EWC and EBLL do not provide many benefits for incremental class learning but still retain more initial knowledge than incremental training normally would. We see that EWC was the least effective in retaining base knowledge across all datasets and loss functions. We note that the EWC technique is still effective for a smaller number of incremental learning steps. Alternatively, EBLL provided decent base knowledge retention without exemplars and suffered less from catastrophic forgetting over a more significant number of incremental steps. However, the overall best performers for gaining new knowledge and retaining knowledge are iCaRL and our approach, as supported by the plots and reinforced further by the discussion around Table~\ref{inc_test} below. 
 
\subsection{Evaluation of the Impact of Loss on Results}

We can observe from the bold and highlighted results in Table~\ref{inc_test} that the four losses with our approach provided the best performance in retaining base knowledge compared to any other combinations. The exception is Center loss, where iCaRL outperformed, shown by $\Omega_{base}$. We can also observe that iCaRL outperformed all the approaches with the four loss functions when learning new classes shown by $\Omega_{base}$. But overall, the bold and highlighted results of $\Omega_{all}$ show that our method mostly performs better across all loss functions. We observe that different loss functions have different extents of forgetting during incremental learning. The best loss is Angular loss, followed by Contrastive, Triplet, and Center loss. Although the different incremental learning techniques reduced the amount of forgetting, we can still compare the differences in the amount of forgetting for the four different loss functions. The Angular loss combined with our approach is the best performer overall, followed by Contrastive, Triplet, and finally, Center loss. The results align with the expectation that different losses have different impacts on forgetting, even with good incremental learning techniques.

\begin{table*}[htb!]
\setlength{\tabcolsep}{2.5pt}
\caption{Incremental class test's mean average precision (mAP@R) starting from the memorised $\Omega_{base}$ with the new classes $\Omega_{new}$ added and the overall result $\Omega_{all}$. Bold is the best method for $\Omega_*$ per loss. The highlight is the best method for $\Omega_*$ per dataset.}\label{inc_test}
\centering
\resizebox{\textwidth}{!}{%
\begin{tabular}{ll|ccc|ccc|ccc|ccc}
\toprule
&  & 
\multicolumn{3}{c|}{\textbf{EWC}}   & 
\multicolumn{3}{c|}{\textbf{EBLL}}  & 
\multicolumn{3}{c|}{\textbf{iCaRL}} & 
\multicolumn{3}{c}{\textbf{Our appr'ch}}  \\ 
\textbf{\makecell[l]{Loss}} & \textbf{Dataset} & 
$\Omega_{base}$ & $\Omega_{new}$ & $\Omega_{all}$ & 
$\Omega_{base}$ & $\Omega_{new}$ & $\Omega_{all}$ & 
$\Omega_{base}$ & $\Omega_{new}$ & $\Omega_{all}$ & 
$\Omega_{base}$ & $\Omega_{new}$ & $\Omega_{all}$ \\ 
\bottomrule \toprule
\multirow{4}{*}{\textbf{\makecell[l]{Contrast'}}} 
    & CIFAR10   & .29 & .16 & .19 & .38 & .11 & .21 & .36 & \cellcolor{black!20}\textbf{.22} & .24 & \cellcolor{black!20}\textbf{.44} & .19 & \textbf{.31}\\
    & MNIST     & .39 & .59 & .37 & .50 & .47 & .43 & .80 & \textbf{.90} & \cellcolor{black!20}\textbf{.91} & \textbf{.89} & .86 & .85\\
    & Fashion-M' & .53 & .45 & .37 & .57 & .46 & .37 & .66 & \cellcolor{black!20}\textbf{.84} & \textbf{.60} & \textbf{.76} & .73 & \cellcolor{black!20}\textbf{.61}\\ 
    & EMNIST     & .19 & .23 & .16 & .26 & .27 & .22 & .56 & \textbf{.55} & .51 & \textbf{.74} & .54 & \textbf{.69}\\
    \midrule
\multirow{4}{*}{\textbf{\makecell[l]{Angular}}} 
    & CIFAR10    & .31 & .11 & .19 & .36 & .13 & .23 & .39 & \textbf{.20} & .27 & \textbf{.40} & .18 & \cellcolor{black!20}\textbf{.39}\\
    & MNIST       & .53 & .51 & .43 & .76 & .51 & .60 & .88 & \textbf{.85} & \textbf{.87} & \cellcolor{black!20}\textbf{.92} & \textbf{.85} & .85\\
    & Fashion-M'  & .63 & .53 & .40 & .70 & .49 & .45 & .74 & \textbf{.74} & .55 & \cellcolor{black!20}\textbf{.77} & .72 & \textbf{.56}\\ 
    & EMNIST     & .27 & .23 & .21 & .43 & .23 & .30 & .55 & .57 & .47 & \cellcolor{black!20}\textbf{.78} & \cellcolor{black!20}\textbf{.63} & \cellcolor{black!20}\textbf{.70}\\
    \midrule
\multirow{4}{*}{\textbf{\makecell[l]{Triplet}}} 
    & CIFAR10     & .24 & .12 & .16 & .33 & .13 & .21 & .32 & \textbf{.19} & \textbf{.23} & \textbf{.39} & .16 & \textbf{.23}\\
    & MNIST      & .45 & .62 & .50 & .56 & .41 & .47 & .78 & \textbf{.90} & .77 & \textbf{.90} & .84 & \textbf{.84}\\
    & Fashion-M'  & .48 & .48 & .27 & .52 & .42 & .31 & .57 & \textbf{.74} & .49 & \textbf{.67} & .73 & \textbf{.52}\\ 
    & EMNIST     & .21 & .23 & .16 & .26 & .16 & .18 & .38 & .41 & .36 & \textbf{.69} & \textbf{.52} & \textbf{.60}\\
    \midrule
\multirow{4}{*}{\textbf{\makecell[l]{Center}}} 
    & CIFAR10     & .18 & .10 & .13 & .18 & .10 & .14 & \textbf{.27} & \textbf{.16} & .19 & .26 & .15 & \textbf{.26}\\
    & MNIST       & .24 & .53 & .27 & .24 & .63 & .30 & \textbf{.86} & \cellcolor{black!20}\textbf{.91} & \textbf{.85} & .83 & .85 & \textbf{.85}\\
    & Fashion-M' & .28 & .43 & .20 & .38 & .50 & .25 & .61 & \textbf{.76} & .50 & \textbf{.71} & .69 & \textbf{.59}\\  
    & EMNIST      & .10 & .20 & .10 & .11 & .25 & .10 & .55 & \textbf{.59} & .49 & \textbf{.60} & .52 & \textbf{.52}\\
    \midrule
\bottomrule
\end{tabular}
}
\end{table*}

Table~\ref{inc_test} presents the results using Equation~\ref{eqn:Metric} for each model. The values: $\Omega_{base}$, $\Omega_{new}$, and $\Omega_{all}$ range between $[0,1]$. $0$ indicates the model retains no knowledge, and $1$ indicates it retains all knowledge. The $\Omega_{new}$ results show the mAP@R performance on test data of the newly learned class. The $\Omega_{all}$ shows how well the models retain prior knowledge and acquire new knowledge. The $\Omega_{new}$ results show that the normal models are learning new knowledge at a very low rate, which would not be useful. In Table~\ref{inc_test}, we evaluated how methods retained previously and newly learnt knowledge by testing on the base-test set (old learned classes) and inc-test set (newly learned classes). The results are standardised with the offline models' ideal performance using Equation~\ref{eqn:Metric}. The offline ideal models' performances were obtained in the same way described earlier, with the difference being we measure the mAP@R on the entire test set (base-test set and inc-test sets combined).

The results in Table~\ref{inc_test} show our approach as the most robust over a long period of incremental class learning, as highlighted by the EMNIST Angular results. Since we are not required to use actual images as exemplars, we can still represent a class well during incremental learning. VAEs are noisy. Despite the noise, they can still represent previously learned classes well. This ability to represent previous classes is important as similarity learning loss functions prioritise separating classes from each other and forming regions for classes to occupy in the embedding space. It is vital to provide the model with information regarding previously learnt classes that occupy existing regions in the embedding space. Table~\ref{inc_test} shows that methods that do not preserve some form of information about previously learnt knowledge are more adversely impacted by catastrophic forgetting.

We observe some interesting differences between iCaRL and our method. Our approach is better than iCaRL overall in terms of overall knowledge retention, but iCaRL is better at learning new classes. Overall, $\Omega_{all}$ shows our approach has better on average mAP@R across all learnt classes. This further supports that we are not required to have images as exemplars to represent previously learnt classes to preserve previous knowledge's embedding spaces. We can represent images in the form of a representation that can be passed through intermediate layers and get similar or better performance than iCaRL. However, we can observe through the CIFAR10 results that simple VAEs might not represent detailed images on more complicated datasets.

Finally, in Table~\ref{inc_test}, we see that Angular loss retains the most base knowledge followed by the Contrastive, Triplet, and Center loss as shown by $\Omega_{base}$ value. Again we note that loss functions with correct mining perform similarly to each other in contrast to previous results~\cite{9311580}. This reinforces the importance of good miners for similarity learning loss functions.

In the following Section, we explore if these results might be attributable to the network architecture rather than the methods and losses.

\subsection{Evaluation of the Impact of Architecture on Results}

\begin{table*}[htb!]
\setlength{\tabcolsep}{2.5pt}
\caption{Incremental class test's mean average precision (mAP@R) starting from the memorised $\Omega_{base}$ with the new classes $\Omega_{new}$ added and the overall result $\Omega_{all}$. Bold is the best method for $\Omega_*$ per loss. Highlight is best method for $\Omega_*$.}\label{vgg11_mnist}
\centering
\resizebox{\textwidth}{!}{%
\begin{tabular}{ll|ccc|ccc|ccc|ccc}
\toprule
& \multicolumn{12}{c}{MNIST}     \\
\hline\hline
\toprule
&  & 
\multicolumn{3}{c|}{\textbf{EWC}}   & 
\multicolumn{3}{c|}{\textbf{EBLL}}  & 
\multicolumn{3}{c|}{\textbf{iCaRL}} & 
\multicolumn{3}{c}{\textbf{Our appr'ch}}  \\ 
\textbf{Architecture} & \textbf{Loss} & 
$\Omega_{base}$ & $\Omega_{new}$ & $\Omega_{all}$ & 
$\Omega_{base}$ & $\Omega_{new}$ & $\Omega_{all}$ & 
$\Omega_{base}$ & $\Omega_{new}$ & $\Omega_{all}$ & 
$\Omega_{base}$ & $\Omega_{new}$ & $\Omega_{all}$ \\ 
\bottomrule \toprule
\multirow{4}{*}{\textbf{VGG11}} 
    & Contrast' &  .42 &  .39 & .42  & .50 & .36 & .41 & \bf{.84} & \bf{.91} & \bf{.81} & .83 & .83 & .80 \\
    & Angular   &  .38 & .43 & .36 & .74 & .37 & .53  & .74 & \bf{.84} &.71 & \bf{.77} & .81 &   \bf{.73}\\
    & Triplet   &   .52 & .55 & .48 & .50 & .36 & .42  & .82 & \bf{.80} & \bf{.78} & \bf{.83} & .80 & .78\\
    & Center   &  .29 & .61 & .32  & .27 & .53 &  .30 & .83 & \cb{.94} & .82 & \cb{.86} & .88 & \cb{.82} \\
\bottomrule
\end{tabular}
}
\end{table*}

\begin{table*}[htb!]
\setlength{\tabcolsep}{2.5pt}
\caption{Incremental class test's mean average precision (mAP@R) starting from the memorised $\Omega_{base}$ with the new classes $\Omega_{new}$ added and the overall result $\Omega_{all}$. Bold is the best method for $\Omega_*$ per loss. Highlight is best method for $\Omega_*$.}\label{vgg11_fmnist}
\centering
\resizebox{\textwidth}{!}{%
\begin{tabular}{ll|ccc|ccc|ccc|ccc}
\toprule
& \multicolumn{12}{c}{FMNIST}     \\
\hline\hline
\toprule
&  & 
\multicolumn{3}{c|}{\textbf{EWC}}   & 
\multicolumn{3}{c|}{\textbf{EBLL}}  & 
\multicolumn{3}{c|}{\textbf{iCaRL}} & 
\multicolumn{3}{c}{\textbf{Our appr'ch}}  \\ 
\textbf{Architecture} & \textbf{Loss} & 
$\Omega_{base}$ & $\Omega_{new}$ & $\Omega_{all}$ & 
$\Omega_{base}$ & $\Omega_{new}$ & $\Omega_{all}$ & 
$\Omega_{base}$ & $\Omega_{new}$ & $\Omega_{all}$ & 
$\Omega_{base}$ & $\Omega_{new}$ & $\Omega_{all}$ \\ 
\bottomrule \toprule
\multirow{4}{*}{\textbf{VGG11}} 
    & Contrast' &   .55 & .34 & .41 & .57 & .26 & .39 & \bf{.76} & \cb{.65} & \cb{.67} &  .76 & .56 & .63\\ 
    & Angular   &   .50 & .32 & .35  &  .64 & .32  & .43  & .65 & .57 & .55 & \cb{.78} & \bf{.59}& \bf{.62}\\ 
    & Triplet   &   .47  &  .36 & .35  & .48 & .32 & .33 & .56 & \bf{.61} & .50 &  \bf{.72} &  .57 & \bf{.58}  \\ 
    & Center    &   .38 & .37 & .29 & .32 & .28 & .23 & .68 & \bf{.65} & \bf{.60} & \bf{.72}  & .58 & .59 \\ 
\bottomrule
\end{tabular}
}
\end{table*}

\begin{table*}[htb!]
\setlength{\tabcolsep}{2.5pt}
\caption{Incremental class test's mean average precision (mAP@R) starting from the memorised $\Omega_{base}$ with the new classes $\Omega_{new}$ added and the overall result $\Omega_{all}$. Bold is the best method for $\Omega_*$ per loss. Highlight is best method for $\Omega_*$.}\label{vgg11_emnist}
\centering
\resizebox{\textwidth}{!}{%
\begin{tabular}{ll|ccc|ccc|ccc|ccc}
\toprule
& \multicolumn{12}{c}{EMNIST}     \\
\hline\hline
\toprule
&  & 
\multicolumn{3}{c|}{\textbf{EWC}}   & 
\multicolumn{3}{c|}{\textbf{EBLL}}  & 
\multicolumn{3}{c|}{\textbf{iCaRL}} & 
\multicolumn{3}{c}{\textbf{Our appr'ch}}  \\ 
\textbf{Architecture} & \textbf{Loss} & 
$\Omega_{base}$ & $\Omega_{new}$ & $\Omega_{all}$ & 
$\Omega_{base}$ & $\Omega_{new}$ & $\Omega_{all}$ & 
$\Omega_{base}$ & $\Omega_{new}$ & $\Omega_{all}$ & 
$\Omega_{base}$ & $\Omega_{new}$ & $\Omega_{all}$ \\ 
\bottomrule \toprule
\multirow{4}{*}{\textbf{VGG11}} 
    & Contrast'  &  .14 & .14  & .13& .23 & .22 & .20 & .60 & \bf{.78}  & .60 & \cb{.81} & .63  & \bf{.72} \\
    & Angular   &  .15 & .25 & .15 & .34 & .25 & .28 & .69 & \bf{.73} &  .62 & \bf{.80}  & .70 & \bf{.68} \\
    & Triplet   &  .14 & .29 & .15 & .20 & .25 & .18 & .56 & .48 & .50 & \bf{.77}  & \bf{.65} & \bf{.66} \\
    & Center    &  .11 & .40 & .12  & .11 & .35 & .12 & \bf{.74} & \cb{.82} & .73  & .74 & .80 & \cb{.74} \\
\bottomrule
\end{tabular}
}
\end{table*}

\begin{table*}[htb!]
\setlength{\tabcolsep}{2.5pt}
\caption{Incremental class test's mean average precision (mAP@R) starting from the memorised $\Omega_{base}$ with the new classes $\Omega_{new}$ added and the overall result $\Omega_{all}$. Bold is the best method for $\Omega_*$ per loss. Highlight is best method for $\Omega_*$.}\label{inc_test_cifar}
\centering
\resizebox{\textwidth}{!}{%
\begin{tabular}{ll|ccc|ccc|ccc|ccc}
\toprule
& \multicolumn{12}{c}{CIFAR10}     \\
\hline\hline
\toprule
&  & 
\multicolumn{3}{c|}{\textbf{EWC}}   & 
\multicolumn{3}{c|}{\textbf{EBLL}}  & 
\multicolumn{3}{c|}{\textbf{iCaRL}} & 
\multicolumn{3}{c}{\textbf{Our appr'ch}}  \\ 
\textbf{Architecture} & \textbf{Loss} & 
$\Omega_{base}$ & $\Omega_{new}$ & $\Omega_{all}$ & 
$\Omega_{base}$ & $\Omega_{new}$ & $\Omega_{all}$ & 
$\Omega_{base}$ & $\Omega_{new}$ & $\Omega_{all}$ & 
$\Omega_{base}$ & $\Omega_{new}$ & $\Omega_{all}$ \\ 
\bottomrule \toprule
\multirow{4}{*}{\textbf{ResNet9}} 
    & Contrast' & .25 &    .13 & .15 & .35 & .12 & .20 & .30 & \bf.14 & .19 & \bf.68 & .05 & \cb.36 \\
    & Angular   & .41 &    .12 & .24 & .43 & .12 & .26 & .34 & \cb.13 & .21 & \cb.74 & .07 & \bf.36 \\
    & Triplet   & .26 &    .12 & .16 & .34 & .12 & .20 & .35 & \bf.13 & .21 & \bf.57 & .07 & \bf.27 \\
    & Center    & .23 & \bf.11 & .14 & .20 & .11 & .10 & .22 & \bf.11 & .14 & \bf.47 & .05 & \bf.24 \\
\bottomrule
\end{tabular}
}
\end{table*}

The results in Table~\ref{inc_test} motivate us to investigate deeper neural network architectures further. To assess if the network architecture might factor in the results, we select ResNet9 as representative of an alternative deep neural network architecture. We select the CIFAR10 dataset as we have previously observed this to be the worst-performing dataset for our technique. We repeat the above experimental setup once with the same metrics using CIFAR10 and a ResNet9. In Table~\ref{inc_test_cifar}, we note that the results are similar to those in Table~\ref{inc_test}, with equivalent techniques outperforming for the same metric. Our method remains the best overall performing technique. It is worth noting that the results are normalised with respect to a baseline ideal and are not directly comparable with Table~\ref{inc_test} due to ResNet9 achieving a higher baseline ideal compared to the simpler CNN.

In Table~\ref{inc_test_cifar}, we see that the ordering of methods other than ours has changed compared to Table~\ref{inc_test_cifar}. There are significant differences between each class of CIFAR10, and base convolution knowledge learnt may not necessarily transfer well to new classes during incremental learning. We speculate that when it comes to learning new classes, the frozen convolution layers of our approach limit its ability to learn new features required to differentiate between new and old classes. The results highlight future research possibilities to improve our technique using VAEs to generate inputs into the convolution layers.

Further, we observe that the ResNet9 architecture improves most methods' ability to retain base knowledge at the cost of new knowledge. We select the VGG11 as additional representative deep neural network architecture. We repeat the above experiments on MNIST, FMNIST, and EMNIST with VGG11. We omitted CIFAR10 as we have already conducted the experiment using ResNet9, as shown in Table~\ref{inc_test_cifar}. The result in Table \ref{vgg11_mnist}, \ref{vgg11_fmnist}, and \ref{vgg11_emnist} show that our method remains the best overall using the VGG11 architecture. However, iCaRL outperforms our method in some situations, which are in line with expectations compared to the results in Table~\ref{inc_test}.

\section{Conclusions}

We investigated to what extent similarity-based loss functions were affected by catastrophic forgetting during incremental similarity learning. We do this by implementing a minor variation of the existing catastrophic forgetting testing procedure and testing metric by Kemker \emph{\textit{et al.}} \cite{kemker2018measuring} to accommodate testing similarity learning. The results show that the extent of catastrophic forgetting differed for each loss function. With good Pair and Triplet mining approaches, we observed that Angular loss was the least affected amongst the loss functions by a small margin. The Centre loss was the most affected. We have found that the severity and rapidness of forgetting depend on the complexity of data (i.e. the number of classes, the task's difficulty) in similarity-based loss functions. Therefore, we have shown that retrieval models with similarity-based loss functions are unsuitable for incremental learning without some form of modification to the training procedure. 

To address the above, we implemented three existing incremental learning methods to reduce catastrophic forgetting during incremental similarity learning and compared them to our novel approach. The results have shown that all the methods helped reduce forgetting in the networks compared to regular training. Our method outperformed the other three methods when measured on both base knowledge retention and overall knowledge retention. We differentiate our method from traditional exemplar methods that require one to keep images. Since using VAEs to generate actual images is a complicated process, we proposed using VAEs to generate intermediate representations instead. The results show that we do not require actual images as exemplars during incremental learning. Instead, it is essential to remind the network of the previously seen knowledge.

The results indicate that ResNet9 outperformed a simple CNN and VGG11 when dealing with catastrophic forgetting. This outcome is exciting as it points to a possible approach of study for overcoming catastrophic forgetting in similarity learning: network architecture. It is also worth noting that the similarity loss functions impact catastrophic forgetting and that, if appropriately formulated, might alleviate some of the challenges. Further research might also investigate the impact of appropriate mining methods in helping prevent catastrophic forgetting. Finally, research into using VAEs or sophisticated generative models to generate inputs to intermediate convolution layers instead of only fully connected ones may be valuable.

Other possible future applications for investigation include applying incremental learning to large language models. Large language models require a significant amount of training data. To update a language model, old and new training data is required resulting in overly long training times. Overcoming catastrophic forgetting would allow one to  update these models rather than retraining from scratch. Generative models are another area for future exploration related to incremental learning. Generative models cannot generate classes of objects that have not been seen before. Incremental learning techniques, overcoming catastrophic forgetting, can be utilized to update the generative models on new classes and not retrain entirely.  

We hope this research sheds insight into the problem of incremental learning with similarity learning. 

\section*{Acknowledgements}
We want to acknowledge the Nedbank Research and Innovation Chair for supporting this research. The financial assistance of the National Research Foundation (NRF) towards this research is hereby acknowledged. Opinions expressed and conclusions arrived at are those of the author and are not necessarily to be attributed to the NRF. 

\section*{Conflict of interest}
 The authors declare no conflicts of interest.
 
\section*{Data Availability}
The datasets analysed during the current study are available in the following repositories: MNIST - \href{https://deepai.org/dataset/mnist}{Link}, FashionMNIST - \href{https://github.com/zalandoresearch/fashion-mnist}{Link}, EMNIST - \href{https://www.westernsydney.edu.au/icns/reproducible_research/publication_support_materials/emnist}{Link} and CIFAR10 - \href{https://www.cs.toronto.edu/~kriz/cifar.html}{Link}. 

\bibliographystyle{spmpsci.bst}      
\bibliography{annot}   

\end{document}